\documentclass[11pt]{article}

\usepackage[final]{acl}

\usepackage{times}
\usepackage{latexsym}

\usepackage{array}
\usepackage{booktabs}
\usepackage{tikz}
\usepackage{multirow}

\usepackage[T1]{fontenc}

\usepackage[utf8]{inputenc}

\usepackage{microtype}

\usepackage{inconsolata}

\usepackage{graphicx}

\usepackage{tcolorbox}
\usepackage{enumitem}
\usepackage{amsmath}
\usepackage{url}
\usepackage{cleveref}
\usepackage{algorithm}
\usepackage{algpseudocode}
\usepackage{amssymb}

\usepackage{tabularx}
\usetikzlibrary{arrows.meta,positioning}

\tcbuselibrary{breakable}
\usepackage{needspace}
\usepackage{ragged2e}
\usepackage{xcolor}
\usepackage{multicol}

\usepackage{tikz}
\usetikzlibrary{positioning,arrows.meta}

\newtcolorbox{cfpanel}[1]{
  breakable,
  width=\columnwidth,
  colback=white,
  colframe=black!22,
  boxrule=.45pt,
  arc=1mm,
  left=4pt,
  right=4pt,
  top=4pt,
  bottom=4pt,
  before skip=6pt,
  after skip=6pt,
  before=\par\Needspace{8\baselineskip},
  fontupper=\footnotesize\RaggedRight,
  title={#1},
  fonttitle=\bfseries\footnotesize,
  colbacktitle=black!5,
  coltitle=black,
  pad at break=1mm
}

\newcommand{\rubcritfull}[8]{%
  \par\noindent\textbf{#1. #2}\par
  \noindent\textit{Description.} #3\par
  \noindent\textbf{1--2:} #4\par
  \noindent\textbf{3--4:} #5\par
  \noindent\textbf{5--6:} #6\par
  \noindent\textbf{7--8:} #7\par
  \noindent\textbf{9--10:} #8\par
  \medskip
}

\newcommand{\rubscore}[4]{%
  \par\noindent\textbf{Criterion #1: #2.}
  \textbf{Score: #3.}
  #4\par\medskip
}

\definecolor{failureorange}{HTML}{E67E22}

\newcommand{\gradientcellfailure}[1]{%
    \begin{tikzpicture}[baseline]
        \fill[failureorange!18] (0,0) rectangle (1.5,0.3);
        \fill[failureorange!95] (0,0) rectangle ({1.5*#1/100},0.3);
        \node[anchor=west, font=\scriptsize] at (1.55,0.15) {
            \textbf{\makebox[2.1em][r]{#1}}
        };
    \end{tikzpicture}%
}

\title{The Illusion of \textit{What If}: Evaluating the Breakdown of \\ Counterfactual Reasoning in LLMs}

\author{\normalfont Yucheng~Wang$^{\spadesuit}$\thanks{Equal contribution.},~Yuetian~Du$^{\spadesuit}$\footnotemark[1],~Zhengyi~Liu$^{\spadesuit}$\footnotemark[1]\thanks{Work done during an internship at Zhejiang University.},~Rongyu~Zhang$^{\spadesuit}$,~Bing~Zhao$^\diamondsuit$ \\  Boyu~Yang$^\diamondsuit$,~Ming~Kong$^\spadesuit$,~Lin~Qu$^\diamondsuit$,~Hu~Wei$^\diamondsuit$\thanks{Corresponding Authors.},~Jie~Liu$^{\clubsuit}$\footnotemark[3],~Qiang~Zhu$^{\spadesuit}$\footnotemark[3] \\[2mm]
$^\spadesuit$ Zhejiang University~~~~$^\diamondsuit$ Alibaba Group~~~~$^\clubsuit$ City University of Hong Kong\\ [1mm]
12621204@zju.edu.cn\quad zhuq@zju.edu.cn
}

\begin{document}
\maketitle
\begin{abstract}
Counterfactual reasoning requires models to reason beyond the observed world and explain how altered conditions propagate through downstream consequences. Existing benchmarks largely target bounded settings with fixed variables or single gold outcomes, overlooking open-domain scenarios requiring causal-process evaluation. To this end, we present \textbf{WhatIfBench}, a diagnostic benchmark for open-domain, open-form, long-horizon counterfactual causal reasoning, containing 220 what-if questions across STEM, HSS, and Hybrid scenarios. To evaluate free-form responses, we further propose \textbf{PRISM}, which first converts each natural-language explanation into a Response-Derived Semantic Causal Graph of events, states, and mechanisms. On top of this graph, PRISM then jointly applies a Process Metric assessing graph-level causal validity and a Rubric Metric assessing answer-level explanatory adequacy. Evaluating six frontier LLMs with this framework, we find that WhatIfBench remains far from saturated: even the strongest model reaches only a 64.62\% final score. Further analysis reveals persistent causal gaps, premise drift, and topology fragmentation, suggesting that fluent counterfactual narratives often mask fragile causal processes. The benchmark, code, and evaluation scripts are available at \url{https://github.com/zju-gt/WhatIfBench}.
\end{abstract}

\section{Introduction}

\begin{tcolorbox}[colback=gray!5, colframe=gray!40, boxrule=0.5pt]

\textit{\small ``We think of a cause as something that makes a difference, and the difference it makes must be a difference from what would have happened without it.''}

\vspace{0.8em}
\small \hfill --- David Lewis, \textit{Causation}
\end{tcolorbox}

\begin{figure}[t!]
    \centering
    \includegraphics[width=0.9\linewidth]{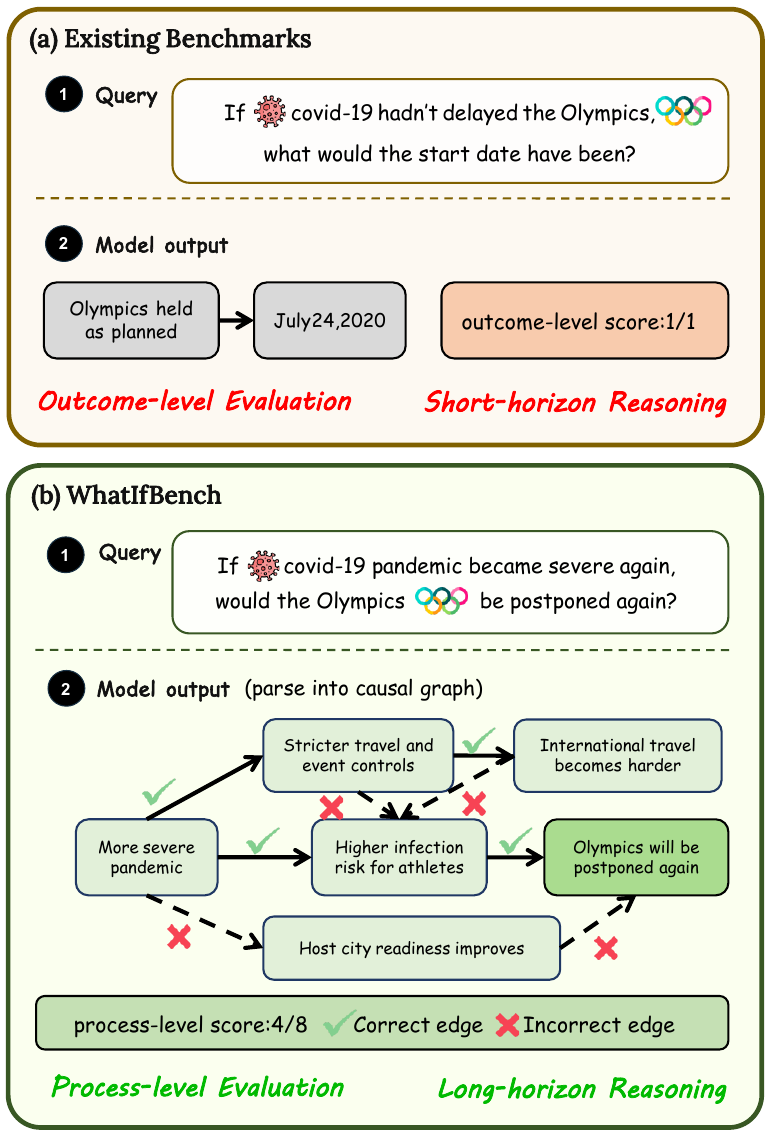}
\caption{
\textbf{Comparison of what-if reasoning evaluation paradigms.}
(a) Existing benchmarks focus on outcome-level evaluation for short-horizon reasoning.
(b) WhatIfBench shifts the focus to process-level evaluation for long-horizon reasoning by parsing model outputs into causal graphs and checking causal edges.
}
    \label{fig:fig1}
\end{figure}

Large language models (LLMs) are increasingly expected to reason not only about what is, but about what would happen under different conditions. Such \textit{what-if} questions arise in policy~\citep{he2026thinking}, science~\citep{zheng2025newtonbench}, history~\citep{nguyen2025counterfactual}, and technology~\citep{vashishtha2025executable}, and require more than factual recall~\citep{bondarenko2022causalqa,ho2022wikiwhy}. They require a model to depart from the observed timeline, identify which factors make a difference to an outcome and propagate their consequences through interacting systems~\citep{zhang2023causal}. They thus serve as a stress test of whether a model's explanations reflect coherent causal processes rather than plausible-sounding narratives.

Existing counterfactual benchmarks test useful but bounded forms of hypothetical reasoning. CRASS converts counterfactual conditionals into QA instances~\citep{frohberg2022crass}, IfQA studies open-domain QA with counterfactual presuppositions~\citep{yu2023ifqa}, and recent formal benchmarks evaluate intervention-style reasoning over specified variables, causal structures, and target outcomes~\citep{jin2023cladder,wang2024causalbench,chen2025counterbench}. These benchmarks test whether models can recognize a counterfactual premise, apply it to a local question, or reason within an explicitly defined causal setup, as shown in Figure~\ref{fig:fig1}. However, many real-world \textit{what-if} questions are less bounded. The mechanisms that matter are often only implicit, consequences may unfold over long horizons, and a single intervention can propagate through social, technical, political, or scientific systems in multiple plausible ways~\citep{tetlock1997counterfactual}. This motivates an open-domain, open-form benchmark defined by broad premises rather than predefined variables or single gold outcomes.

This setting raises a central evaluation problem: \textit{when no unique gold outcome exists, what should be judged?} Under the premise ``What if smartphones had never been widely adopted?'', for example, valid answers may emphasize slower mobile Internet diffusion, alternative portable devices, or different patterns of social coordination. The key issue is not whether these answers converge to the same outcome, but whether each remains faithful to the premise and supports its consequences through a coherent causal structure. A plausible conclusion can still be unreliable if it skips mechanisms, reverses causal direction, or reverts to factual-world assumptions. Following counterfactual accounts of possible alternatives~\citep{lewis1973causation} and interventionist views of causal explanation~\citep{woodward2005making}, we argue that evaluation should target the causal explanation rather than the final outcome alone. A benchmark should therefore make free-form answers inspectable, and evaluate premise fidelity, mechanism coverage, domain consistency, and uncertainty handling.

Based on this principle, we introduce \textbf{WhatIfBench}, a diagnostic benchmark for counterfactual causal reasoning in open-domain, long-horizon scenarios. It contains 220 open-form what-if questions across STEM (science, technology, engineering, and mathematics), HSS (humanities and social sciences), and Hybrid scenarios. STEM questions focus on natural and technical constraints, HSS questions focus on historical, institutional, and social dynamics, and Hybrid questions involve causal propagation across domains. For each query, WhatIfBench collects model responses as open-form causal explanations, preserving the natural form of counterfactual reasoning while setting up structured evaluation.

To evaluate these explanations, we further introduce \textbf{PRISM} (\textbf{P}rocess-and-\textbf{R}ubric \textbf{I}ntegrated \textbf{S}coring \textbf{M}echanism), an evaluation framework for structured assessment of these explanations. Specifically, PRISM first converts each natural-language response into a \textbf{Response-Derived Semantic Causal Graph} by extracting events, states, and mechanisms and organizing them into a causal DAG (directed acyclic graph). On top of this graph, \textbf{Process Metric} (PM) evaluates graph-level causal validity, including edge validity, causal direction, mechanism support, and topology continuity. In parallel, \textbf{Rubric Metric} (RM) evaluates answer-level explanatory adequacy with question-specific rubrics, including premise fidelity, key mechanism coverage, domain constraints, and uncertainty handling. Together, PM and RM provide a fine-grained diagnostic view of open-domain counterfactual reasoning beyond fixed-answer matching.

Our contributions are summarized as follows:
\begin{itemize}[leftmargin=*, itemsep=0pt, topsep=0pt, parsep=1pt]
    \item \textbf{WhatIfBench.} We introduce a diagnostic benchmark for counterfactual causal reasoning in open-domain, open-form, long-horizon scenarios, containing 220 open-form what-if questions across STEM, HSS, and Hybrid scenarios.
    
    \item \textbf{Response-Derived Semantic Causal Graphs.} We convert natural-language explanations into response-derived semantic causal graphs, making the causal structure expressed in model responses inspectable without requiring graph-form outputs.
    
    \item \textbf{PRISM Evaluation.} We develop \textbf{PRISM}, a process-and-rubric integrated evaluation framework that jointly assesses graph-level causal validity and fine-grained answer-level explanatory adequacy.
\end{itemize}

\section{Related Work}

\noindent\textbf{Causal reasoning in LLMs.}
Causal reasoning is a key capability for evaluating whether LLMs can go beyond surface associations and produce robust explanations of how events depend on one another~\citep{kiciman2023causal,yu2025causaleval,wu2024causality,jin2024can,miliani2025explica}. Although LLMs often perform well on causal tasks, it remains unclear whether this reflects genuine causal understanding or parametric associations from pretraining~\citep{chi2024unveiling,yang2024critical,yamin2024failure,wang2026causalflip}. Counterfactual reasoning sharpens this issue: models must depart from the factual world, reason under a hypothetical intervention, and propagate its consequences, yet they often revert to factual-world assumptions when counterfactual premises conflict with parametric knowledge~\citep{yamin2025can,balappanawar2025if}. We study this capability in open-domain, long-horizon scenarios, where success requires coherent causal processes rather than plausible short answers.

\noindent\textbf{Benchmarks for counterfactual reasoning.}
Existing counterfactual reasoning benchmarks have begun to test LLMs under hypothetical assumptions. CRASS~\citep{frohberg2022crass} evaluates questionized counterfactual conditionals, IfQA~\citep{yu2023ifqa} introduces counterfactual presuppositions into open-domain question answering, and CounterBench~\citep{chen2025counterbench} studies formally specified counterfactual questions with controlled causal structures and difficulty levels. These benchmarks provide valuable diagnostics, but they are often built around short conditionals, answerable QA instances, or formal settings with relatively constrained variables and outcomes~\citep{jin2023cladder,wang2024causalbench,yang2024critical}. As a result, open-form counterfactual reasoning questions remain underexplored, especially when they are underspecified, span interacting systems, and admit multiple plausible trajectories.

\noindent\textbf{Evaluation of counterfactual reasoning.}
The evaluation of counterfactual reasoning remains largely answer-centric~\citep{he2026uncovering}. IfQA~\citep{yu2023ifqa} uses Exact Match and token-level F1 against reference answers, CRASS~\citep{frohberg2022crass} compares predictions with human-validated answers, and CounterBench~\citep{chen2025counterbench} evaluates whether models derive the correct final value or binary outcome under formal counterfactual settings. Such metrics are suitable when a task has a short reasoning path or a verifiable target answer, but they are insufficient when no unique gold outcome exists~\citep{yang2024critical,yu2025causaleval}. In open-domain long-horizon counterfactual reasoning, a response may sound plausible while drifting from the premise, skipping mechanisms, reversing causal direction, or introducing unsupported causal links. This calls for evaluation that exposes the causal process expressed in a response and assesses explanatory adequacy beyond final-answer matching.

\begin{figure*}[t!]
    \centering
    \includegraphics[width=\linewidth]{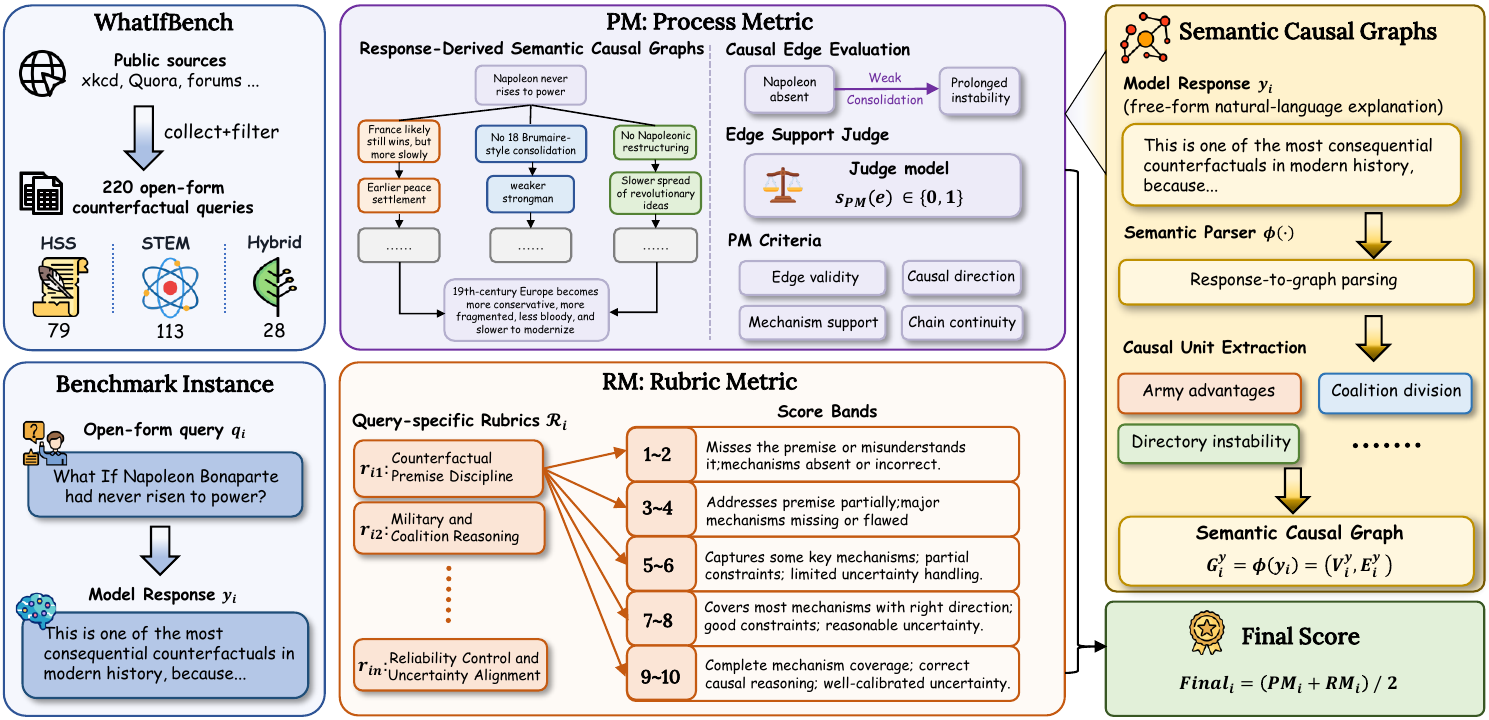}
\caption{
\textbf{Overview of WhatIfBench and PRISM.}
WhatIfBench contains open-form counterfactual queries spanning STEM, HSS, and Hybrid scenarios. For each query, a model generates a free-form response, which PRISM evaluates through two complementary components: the \textbf{Process Metric} (PM), which parses the response into a response-derived semantic causal graph and validates its causal edges, and the \textbf{Rubric Metric} (RM), which scores answer-level explanatory adequacy using query-specific rubrics. The final score combines PM and RM, enabling evaluation of both causal-process validity and answer-level explanatory adequacy.
}
    \label{fig:fig2}
\end{figure*}

\section{WhatIfBench}

\subsection{Task Definition}

We define the task as \textit{open-domain, open-form, long-horizon counterfactual causal reasoning}. Let \(x_i=(q_i,I_i)\) denote a what-if instance, where \(q_i\) specifies the question context and \(I_i\) specifies an explicit counterfactual premise or intervention. Given \(x_i\), a model \(M\) generates a free-form natural-language explanation:
\begin{equation}
    x_i \xrightarrow{M} y_i.
\end{equation}

The generated explanation \(y_i\) is expected to remain faithful to the premise \(I_i\), instantiate relevant intermediate mechanisms, and propagate downstream consequences from the counterfactual condition. The task has three defining properties. 

\noindent\textit{1) Open-domain} indicates that instances span broad domains with underspecified causal spaces. 

\noindent\textit{2) Open-form} indicates that outputs are natural-language explanations rather than predefined choices or constrained spans. 

\noindent\textit{3) Long-horizon} indicates that reasoning should cover multi-step causal propagation across nontrivial causal structures.

\subsection{Benchmark Overview}

WhatIfBench contains \textbf{220 open-form what-if questions} across \textbf{113 STEM}, \textbf{79 HSS}, and \textbf{28 Hybrid} scenarios. Each instance consists of a normalized query \(q_i\), an explicit counterfactual premise \(I_i\), a domain category \(d_i\), and a question-level rubric \(\mathcal{R}_i\). When background materials or reference discussions are available, we use them to identify plausible constraints, mechanisms, and explanatory requirements, without assigning a unique gold conclusion.

\subsection{Taxonomy of Counterfactual Scenarios}

We categorize each question into one of three scenario types:

\noindent\textbf{STEM} covers scientific and technical mechanisms governed by relatively stable constraints, such as physical laws, biological mechanisms, engineering systems, and technological dependencies, testing whether models can propagate counterfactual consequences without violating domain constraints.

\noindent\textbf{HSS} covers historical, institutional, social, cultural, and geopolitical dynamics, emphasizing path dependence, multi-agent behavior, social feedback, and long-term institutional consequences.

\noindent\textbf{Hybrid} covers cross-domain scenarios where natural, technological, environmental, or resource changes propagate into economic, social, or institutional systems, testing whether models can maintain coherent causal topology across heterogeneous domains.

This taxonomy supports both overall model comparison and category-level diagnosis, allowing us to distinguish failures caused by scientific constraint violations, weak socio-historical reasoning, or cross-domain propagation errors.

\subsection{Data Collection and Annotation}

We construct WhatIfBench through candidate collection, filtering, annotation, and review. Candidate questions are gathered from public what-if QA sites, online discussion forums, and explanatory materials, including \textit{xkcd What If}\footnote{\url{https://what-if.xkcd.com/}}, \textit{Worldbuilding Stack Exchange}\footnote{\url{https://worldbuilding.stackexchange.com/}}, \textit{AlternateHistory}\footnote{\url{https://www.alternatehistory.com/}}, and \textit{Quora}\footnote{\url{https://www.quora.com/}}. We rewrite retained questions into normalized queries and keep only those with an explicit counterfactual premise, sufficient contextual grounding, and nontrivial causal propagation requirements. For each retained question, we annotate the domain category \(d_i\), summarize the plausible reasoning space, and construct a query-dependent rubric \(\mathcal{R}_i\).

\subsection{Quality Control}

We conduct quality control at both the question and rubric levels. Question-level review checks whether each example has a clear counterfactual premise, sufficient contextual grounding, nontrivial causal propagation requirements, and an evaluable explanatory structure. Rubric-level review checks whether the criteria cover the core reasoning space while allowing multiple plausible downstream trajectories. As shown in Table~\ref{tab:quality_control}, 36.0\% of 611 candidate questions are retained after direct pass or revision, yielding 220 final questions. All corresponding rubric sets pass after direct review or revision.

\begin{table}[t]
\centering
\small
\setlength{\tabcolsep}{6pt}
\begin{tabular}{@{}lccccc@{}}
\toprule
\textbf{Target} & \textbf{\#} & \textbf{Pass} & \textbf{Revised} & \textbf{Rejected} & \textbf{Valid} \\
\midrule
Questions & 611 & 14.9 & 21.1 & 64.0 & 36.0 \\
Rubrics   & 220 & 92.3 & 7.7  & 0.0  & 100.0 \\
\bottomrule
\end{tabular}
\caption{\textbf{Quality control results for WhatIfBench.} All values except \# are percentages at the audit-target level. Valid denotes Pass + Revised.}
\label{tab:quality_control}
\end{table}

\section{Evaluation Framework: PRISM}

We introduce \textbf{PRISM}, a \textbf{P}rocess-and-\textbf{R}ubric \textbf{I}ntegrated \textbf{S}coring \textbf{M}echanism for evaluating open-form counterfactual explanations. As shown in Figure~\ref{fig:fig2}, PRISM produces two scores for each response: \textbf{Process Metric (PM)}, which evaluates causal-edge validity in a response-derived graph, and \textbf{Rubric Metric (RM)}, which evaluates answer-level explanatory adequacy with frozen question-specific rubric set.

\subsection{Response-Derived Semantic Causal Graph}

Given a question \(q_i\), counterfactual premise \(I_i\), and model response \(y_i\), PRISM parses the response into a response-derived semantic causal graph:
\begin{equation}
    G_i^y=\phi(y_i;q_i,I_i)=(V_i^y,E_i^y).
\end{equation}
Here, \(V_i^y\) denotes elementary discourse units or response-level claims segmented from \(y_i\), and \(E_i^y\) denotes directed semantic relations among them. Each edge is represented as
\begin{equation}
    e=(v_a \rightarrow v_b, r_e),
\end{equation}
where \(v_a\) and \(v_b\) are source and target units, and \(r_e\) is a relation label.

The graph is not intended to be a gold structural causal model of the real world. Instead, it is an evaluation-oriented intermediate representation that makes the causal structure expressed in the response inspectable while allowing models to answer naturally in free text. PRISM constructs \(G_i^y\) with a Rhetorical Structure Theory (RST)-inspired parsing scheme~\citep{mann1987rhetorical}. It segments the response into discourse units, identifies semantic relations, and projects them into a directed graph. Relation labels include discourse relations such as \textsc{Elaboration}, \textsc{Contrast}, \textsc{Temporal}, and \textsc{Background}, and causal-relevant relations such as \textsc{Cause}, \textsc{Condition}, \textsc{Result}, and \textsc{Consequence}.

For process scoring, PRISM uses the causal-relevant edge subset:
\begin{equation}
    E_i^{c}=\{e\in E_i^y \mid r_e\in\mathcal{C}\},
\end{equation}
where \(\mathcal{C}\) includes \textsc{Cause}, \textsc{Condition}, \textsc{Result}, and \textsc{Consequence}. Non-causal discourse edges are retained for inspection but excluded from PM.

\subsection{Process Metric}

PM verifies the local validity of each causal-relevant transition by checking textual support, relation type, causal direction, and mechanism-level plausibility. For each edge \(e=(v_a\rightarrow v_b,r_e)\in E_i^c\), PRISM uses a strict binary judge:
\begin{equation}
s_{\mathrm{PM}}(e)=
\begin{cases}
1, & \text{if } e \text{ is explicitly supported by } y_i,\\
0, & \text{otherwise.}
\end{cases}
\end{equation}

An edge is supported only when both endpoint claims are stated or unambiguously paraphrased in the response, the source-to-target relation is locally recoverable from the text, the relation label matches the expressed dependency, and the direction is not ambiguous, reversed, circular, or contradicted. For causal-relevant labels, the response must express an actual causal, conditional, resultative, or consequential dependency rather than a mere temporal sequence, topical association, or relation inferred from external knowledge.

The PM score is the mean support score over all causal-relevant edges:
\begin{equation}
\mathrm{PM}(y_i)=
\begin{cases}
\frac{1}{|E_i^{c}|}\sum_{e\in E_i^{c}}s_{\mathrm{PM}}(e), & |E_i^{c}|>0,\\
0, & |E_i^{c}|=0.
\end{cases}
\end{equation}
Thus, PM should be interpreted as a strict causal-edge validity score over the response-derived semantic causal graph, rather than as a full judgment of real-world causal correctness.

\begin{table*}[t!]
\centering
\small
\setlength{\tabcolsep}{8pt}
\renewcommand{\arraystretch}{1.12}
\begin{tabular}{lccccccccc}
\toprule
\multirow{2}{*}{\textbf{Model}} 
& \multicolumn{2}{c}{\textbf{STEM}} 
& \multicolumn{2}{c}{\textbf{HSS}} 
& \multicolumn{2}{c}{\textbf{Hybrid}} 
& \multicolumn{2}{c}{\textbf{Overall}} 
& \multirow{2}{*}{\textbf{Final}} \\
\cmidrule(lr){2-3}
\cmidrule(lr){4-5}
\cmidrule(lr){6-7}
\cmidrule(lr){8-9}
& PM & RM & PM & RM & PM & RM & PM & RM & \\
\midrule
GLM-5.1 
& 57.72 & 52.19 & 49.63 & 46.93 & 52.83 & 49.92 & 54.07 & 49.92 & 51.99 \\

DeepSeek-V4-Pro 
& 64.43 & 55.81 & 49.53 & \underline{52.74} & 39.42 & 54.33 & 55.78 & 54.50 & 55.12 \\

Qwen3-Max 
& 64.92 & 48.32 & 49.42 & 43.17 & 55.45 & 48.72 & 58.07 & 46.45 & 52.26 \\

Gemini-3.1-Pro-Preview 
& 68.37 & 53.75 & 52.28 & 50.61 & 51.24 & 50.72 & 60.43 & 52.24 & 56.33 \\

Claude-Opus-4.7 
& \underline{69.13} & \underline{56.04} & \textbf{57.72} & 52.02 & \underline{56.08} & \underline{58.48} & \underline{63.39} & \underline{54.82} & \underline{59.11} \\

GPT-5.5 
& \textbf{72.77} & \textbf{64.00} & \underline{57.38} & \textbf{61.98} & \textbf{65.32} & \textbf{64.80} & \textbf{65.94} & \textbf{63.31} & \textbf{64.62} \\
\bottomrule
\end{tabular}
\caption{\textbf{Main results on WhatIfBench.}
We report the Process Metric (PM) and Rubric Metric (RM) across STEM, HSS, Hybrid, and overall categories, together with the final aggregated score. All scores are percentages on a 0--100 scale. The best result in each column is shown in \textbf{bold}, and the second-best result is \underline{underlined}.}
\label{tab:main_results}
\end{table*}




\subsection{Rubric Metric}

PRISM further employs a frozen question-level rubric set \(\mathcal{R}_i\) to compute the Rubric Metric (RM), which measures whether the response covers the required mechanisms, constraints, and downstream consequences for the given question.

For each question \(q_i\), its rubric set is defined as
\begin{equation}
    \mathcal{R}_i=\{r_{i1}, r_{i2}, \ldots, r_{ij}, \ldots, r_{in}\},
\end{equation}
where \(n\) is the number of rubric criteria for question \(q_i\), and each \(r_{ij}\) specifies one question-dependent requirement. These criteria cover premise fidelity, mechanism coverage, downstream consequences, domain constraints, uncertainty handling, and explanatory completeness. For each criterion \(r_{ij}\), the evaluator assigns an integer score
\begin{equation}
    S_{\mathrm{RM}}(y_i, r_{ij})\in\{1,\ldots,10\},
\end{equation}
according to the criterion-specific score-band descriptions associated with \(r_{ij}\). The normalized RM score is:
\begin{equation}
    \mathrm{RM}(y_i)=\frac{1}{10n}\sum_{j=1}^{n}S_{\mathrm{RM}}(y_i, r_{ij}).
\end{equation}
The evaluator rewards only evidence explicitly present or unambiguously paraphrased in the answer and penalizes premise drift, missing mechanisms, ignored constraints, unsupported causal leaps, and overconfident speculation. The rubric set does not prescribe a single gold trajectory or conclusion; instead, it specifies question-dependent requirements that any plausible counterfactual explanation should satisfy.

\subsection{Evaluator Standardization}

We use \textit{GPT-5.4} as the unified parser and judge for PRISM. All parsing and judging prompts are fixed across evaluated models, and the same evaluator is applied to every response under identical decoding settings. PM uses binary edge-level judgments, while RM uses criterion-specific score bands. Complete parsing schemas, judging prompts, rubric templates, and decoding settings are provided in Appendices~\ref{app:decoding_settings} and~\ref{app:prompts}.

\subsection{Final Score and Reporting}

After evaluating all responses, we report model-level All PM and All RM by averaging the corresponding PM and RM scores over all examples. PM is local and process-oriented, while RM is answer-level and rubric-oriented. We use equal weighting as a default diagnostic aggregation and report PM/RM separately to avoid hiding trade-offs between process-level causal validity and answer-level explanatory adequacy. The final score is:
\begin{equation}
    \mathrm{Final\ Score}=\frac{\mathrm{All\ PM}+\mathrm{All\ RM}}{2}.
\end{equation}
We report the same metrics for STEM, HSS, and Hybrid subsets to support category-level diagnosis.

\begin{table*}[t]
    \centering
    \small
    \scalebox{0.92}{
    \begin{tabular}{@{}l|c|c|c|c|c@{}}
    \toprule
    \addlinespace[0.7em]
    \textbf{Model Names} &
    \shortstack{\textbf{Premise}\\\textbf{Drift (\%)}} &
    \shortstack{\textbf{Causal}\\\textbf{Gap (\%)}} &
    \shortstack{\textbf{Topology}\\\textbf{Fragmentation (\%)}} &
    \shortstack{\textbf{Mechanism}\\\textbf{Missing (\%)}} &
    \shortstack{\textbf{Rubric}\\\textbf{Misalignment (\%)}} \\
    \addlinespace[0.3em]
    \midrule

    GLM-5.1 & \gradientcellfailure{38.6} & \gradientcellfailure{66.5} & \gradientcellfailure{48.7} & \gradientcellfailure{48.1} & \gradientcellfailure{46.8} \\

    DeepSeek-V4-Pro & \gradientcellfailure{35.5} & \gradientcellfailure{70.3} & \gradientcellfailure{37.0} & \gradientcellfailure{29.7} & \gradientcellfailure{41.3} \\

    Qwen3-Max & \gradientcellfailure{38.8} & \gradientcellfailure{59.9} & \gradientcellfailure{55.3} & \gradientcellfailure{57.2} & \gradientcellfailure{44.7} \\

    Gemini-3.1-Pro-Preview & \gradientcellfailure{38.8} & \gradientcellfailure{61.9} & \gradientcellfailure{47.0} & \gradientcellfailure{47.8} & \gradientcellfailure{54.5} \\

    Claude-Opus-4.7 & \gradientcellfailure{30.3} & \gradientcellfailure{60.6} & \gradientcellfailure{41.3} & \gradientcellfailure{34.9} & \gradientcellfailure{27.5} \\

    GPT-5.5 & \gradientcellfailure{14.5} & \gradientcellfailure{49.3} & \gradientcellfailure{42.0} & \gradientcellfailure{18.8} & \gradientcellfailure{15.9} \\

    \bottomrule
    \end{tabular}
    }
    \caption{
Failure mode distribution among weak WhatIfBench responses.
Percentages are computed over responses with final PRISM score below 60.
Labels are not mutually exclusive.
}
    \label{tab:failure_modes}
\end{table*}

\section{Experiments}

\subsection{Experimental Setup}

\noindent\textbf{Models.}
We evaluate six frontier LLMs on WhatIfBench: GLM-5.1~\citep{glm5team2026glm5vibecodingagentic}, DeepSeek-V4-Pro~\citep{deepseekai2026deepseekv4}, Qwen3-Max~\citep{yang2025qwen3}, Gemini-3.1-Pro-Preview, Claude-Opus-4.7, and GPT-5.5. These models cover recent high-performing systems with strong general reasoning and open-form generation capabilities.

\noindent\textbf{Benchmark and protocol.}
We run each model on the fixed WhatIfBench test set using the same response-generation template. Responses are generated with temperature $\tau=0.6$ and a 4,096-token budget, and are submitted verbatim to the standardized PRISM pipeline, with evaluator-side decoding set to temperature $\tau=0.0$. We do not apply model-specific prompt variants, answer normalization, or post-hoc correction. Results are reported over the STEM, HSS, and Hybrid splits, together with All PM, All RM, and the equal-weighted final score. Detailed prompts and decoding budgets are provided in Appendices~\ref{app:decoding_settings} and~\ref{app:prompts}.

\subsection{Main Results}

Table~\ref{tab:main_results} presents the main results on WhatIfBench. We summarize three key findings below.

\noindent\textbf{Finding 1: WhatIfBench remains challenging for frontier models.}
Even the strongest model, GPT-5.5, achieves a final score of only 64.62, followed by Claude-Opus-4.7 with 59.11. All evaluated models remain far from perfect performance, indicating that open-domain, open-form, long-horizon counterfactual causal reasoning is still challenging for current frontier LLMs. This suggests that fluent generation does not necessarily imply reliable reasoning under counterfactual premises, especially when models must propagate consequences across interacting mechanisms.

\noindent\textbf{Finding 2: Domain shifts reveal different causal weaknesses.}
Models perform consistently better on STEM scenarios, where counterfactual consequences are often governed by more stable scientific or technical mechanisms. HSS scenarios are generally more challenging, reflecting difficulties in modeling path dependence, institutional dynamics, and multi-agent social consequences. Hybrid scenarios are not uniformly the hardest, but they expose stronger model-specific variation and PM--RM trade-offs, suggesting that cross-domain reasoning may cover relevant outcomes without maintaining coherent causal propagation.

\noindent\textbf{Finding 3: Process validity and explanatory adequacy capture complementary abilities.}
Overall PM is higher than overall RM for every model, but the gap varies substantially across systems. For example, Qwen3-Max achieves a relatively strong overall PM of 58.07 but a much lower overall RM of 46.45, suggesting that locally plausible causal transitions do not necessarily yield a complete or well-grounded explanation. The strongest models, GPT-5.5 and Claude-Opus-4.7, rank first and second on both PM and RM, indicating that stronger counterfactual reasoning requires both coherent causal structure and sufficient answer-level coverage. This supports the design of PRISM as a joint evaluation of process-level causal validity and rubric-level explanatory adequacy.

\begin{figure}[t!]
    \centering
    \includegraphics[width=\linewidth]{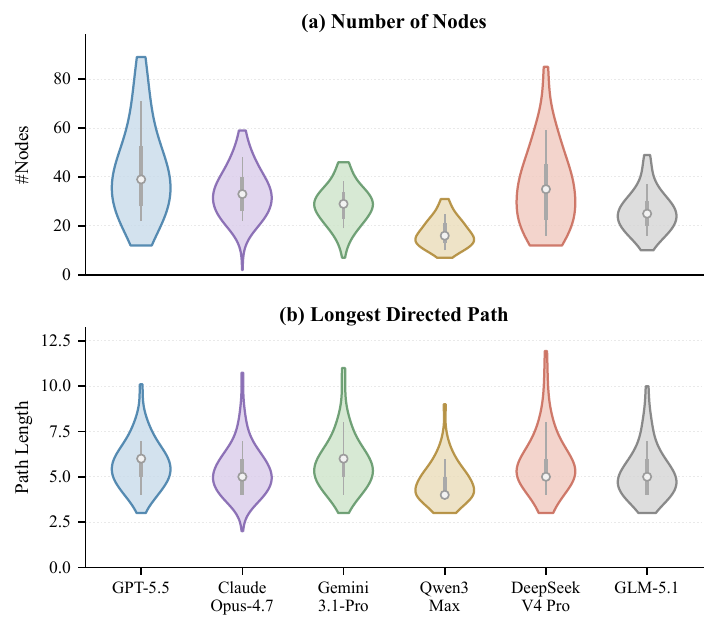}
\caption{
\textbf{Topology of response-derived causal graphs across models.}
We visualize the distributions of two graph-level structural statistics extracted from parsed model responses: 
(a) the number of nodes, measuring the scale of instantiated causal factors, and 
(b) the longest directed path length, measuring the depth of causal propagation.
}
\label{fig:topology_violin}
\end{figure}

\subsection{Failure Mode Analysis}

Aggregate scores identify performance gaps, but they do not reveal which part of counterfactual reasoning breaks. We therefore analyze weak responses with final PRISM score below 60 and assign five non-mutually-exclusive labels using PRISM scores and judge rationales: \textbf{Premise Drift} for deviations from the counterfactual premise, \textbf{Causal Gap} for unsupported causal edges or missing bridges, \textbf{Topology Fragmentation} for broken long-horizon continuity, \textbf{Mechanism Missing} for insufficient process-level detail, and \textbf{Rubric Misalignment} for failures to satisfy question-specific explanatory requirements.

\noindent\textbf{Finding 4: Counterfactual failures are structural rather than merely factual.}
Table~\ref{tab:failure_modes} shows that weak responses are dominated by causal-process failures. \textbf{Causal Gap} is the most frequent failure mode for every model, indicating that models often mention plausible consequences without establishing the intermediate causal bridges from the counterfactual premise. Together with Topology Fragmentation and Mechanism Missing errors, this suggests that many failures arise not because responses are entirely off-topic, but because they fail to sustain coherent, mechanism-grounded causal propagation. This supports the motivation of WhatIfBench and PRISM: open-form counterfactual evaluation must inspect the structure of reasoning, not only the surface adequacy of final answers.

\begin{table}[t]
\centering
\small
\caption{\textbf{Human audit of response-to-graph parsing quality.} All values are percentages. Node, Edge, Graph, and Causal Verification denote the validity of extracted nodes, directed edges, graph structure, and causal relations, respectively.}
\label{tab:parser_audit}

\resizebox{0.47\textwidth}{!}{%
\begin{tabular}{@{}lcccc@{}}
\toprule
\textbf{Audit Set} & \textbf{Node} & \textbf{Edge} & \textbf{Graph} & \textbf{Causal} \\
& \textbf{Validity} & \textbf{Validity} & \textbf{Validity} & \textbf{Verification} \\
\midrule
Human1 & 99.6 & 88.7 & 98.3 & 94.2 \\
Human2 & 100.0 & 92.3 & 98.3 & 93.4 \\
Human3 & 98.7 & 89.5 & 100.0 & 94.7 \\
\bottomrule
\end{tabular}%
}
\end{table}

\begin{table}[t]
\centering
\small
\caption{\textbf{Correlation with human judgments.} H--H denotes Human--Human agreement, and P--H denotes PRISM--Human agreement.}
\label{tab:human_correlation}

\resizebox{0.47\textwidth}{!}{%
\begin{tabular}{@{}llccc@{}}
\toprule
\textbf{Score} & \textbf{Pair} & \textbf{Pearson $r$} & \textbf{Spearman $\rho$} & \textbf{MAE} \\
\midrule
\multirow{2}{*}{PM}
& H--H & 0.918 & 0.910 & 0.075 \\
& P--H & 0.812 & 0.846 & 0.101 \\
\midrule
\multirow{2}{*}{RM}
& H--H & 0.921 & 0.888 & 0.029 \\
& P--H & 0.875 & 0.726 & 0.034 \\
\midrule
\multirow{2}{*}{Final}
& H--H & 0.937 & 0.903 & 0.037 \\
& P--H & 0.833 & 0.815 & 0.059 \\
\bottomrule
\end{tabular}
}
\end{table}

\subsection{Visualization}

\noindent\textbf{Finding 5: Graph scale and causal depth capture different structural behaviors.}
Figure~\ref{fig:topology_violin} shows that broader explanations are not necessarily deeper causal explanations. GPT-5.5 and DeepSeek-V4-Pro tend to produce larger response-derived causal graphs, suggesting broader causal coverage, whereas Gemini-3.1-Pro achieves comparable longest-path depth with fewer nodes. This indicates that models differ not only in how many causal elements they mention, but also in how efficiently they organize them within the response-derived causal graph. The result reinforces the need for structure-aware evaluation: surface-level explanatory richness should not be conflated with coherent causal propagation.

\subsection{Correlation with Human Judgments}

Because PRISM relies on automatic response-to-graph parsing and evaluator judgments, we examine its reliability from two complementary perspectives. We first audit whether the parser can recover the causal process expressed in model responses. We then compare PRISM scores with human judgments under a Human--Human agreement ceiling.

\noindent\textbf{Finding 6: Response-derived causal graphs provide a reliable basis for process evaluation.}
Table~\ref{tab:parser_audit} shows that response-to-graph parsing is stable across three human audit sets. Node Validity remains above 98.7\%, Graph Validity above 98.3\%, and Causal Verification above 93.4\%, indicating that the parser generally recovers valid response-level units, coherent graph structures, and plausible causal relations. Edge Validity is lower, ranging from 88.7\% to 92.3\%, suggesting that causal-edge extraction is the most challenging part of the parsing pipeline. Overall, these results support the use of response-derived semantic causal graphs as a reliable basis for downstream PM evaluation.

\noindent\textbf{Finding 7: PRISM captures most human evaluation signal under a human-agreement ceiling.}
Table~\ref{tab:human_correlation} shows that PRISM aligns strongly with human judgments while remaining below the Human--Human reference. For PM, PRISM achieves 0.812 Pearson and 0.846 Spearman correlation, close to the Human--Human agreement of 0.918 and 0.910. For RM, PRISM obtains a high Pearson correlation of 0.875 and an MAE of 0.034, near the Human--Human MAE of 0.029, although its lower Spearman correlation indicates remaining rank-order differences. These results suggest that PRISM captures the majority of the human evaluation signal under a realistic human-agreement ceiling, supporting its use as a scalable diagnostic evaluator rather than a substitute for human assessment.




\section{Conclusion}


We presented \textbf{WhatIfBench}, a benchmark for open-domain, open-form, long-horizon counterfactual causal reasoning, and \textbf{PRISM}, a process-and-rubric framework for evaluating counterfactual explanations. By converting model responses into response-derived semantic causal graphs and combining process- and rubric-level assessment, PRISM enables diagnosis beyond final-answer matching. Experiments on six frontier LLMs show that WhatIfBench remains far from saturated: models often produce fluent narratives while exhibiting causal gaps, premise drift, and topology fragmentation. These findings suggest that counterfactual reasoning should be evaluated as a causal explanation process, not merely as outcome prediction.

\section*{Acknowledgments}
This work was supported by the National Natural Science Foundation of China under Grant 42394060 and 42394064, and Alibaba Group through Alibaba Research Intern Program.

\section*{Limitations}

Although WhatIfBench supports controlled evaluation of open-domain, open-form, long-horizon counterfactual reasoning, its coverage remains necessarily limited. The benchmark contains 220 questions across STEM, HSS, and Hybrid scenarios, which enables systematic comparison but cannot fully represent the diversity of counterfactual settings, specialized domains, or culturally grounded assumptions. Since the questions are collected and normalized from public English-centric sources, the benchmark may also inherit topical and stylistic biases from these sources. We will extend WhatIfBench with more domains, more questions, and broader linguistic and cultural coverage.

PRISM should also be understood as an automatic diagnostic evaluator rather than a gold-standard causal truth oracle. It evaluates response-derived semantic causal graphs rather than gold structural causal models of the world. Therefore, the Process Metric measures causal coherence, textual support, and structural consistency within a model response, but does not provide a definitive judgment of real-world causal truth. PRISM may also be affected by evaluator bias, parsing errors, and rubric design choices.
Future work may combine counterfactual causal evaluation with fine-grained signals such as model confidence~\citep{du2025confidence,Du_2026_CVPR,du2026care} and trajectory-level metrics~\citep{du2026reveal}, e.g., to detect overconfident causal leaps or to filter high-quality counterfactual reasoning trajectories.
We will further extend the evaluation setting beyond single-turn text-only frontier LLMs to retrieval-augmented, interactive, multimodal, and human-in-the-loop counterfactual reasoning.

\section*{Ethical Considerations}

Counterfactual causal reasoning is a dual-use capability. Evaluating whether a model can reason under an explicitly stated what-if premise is, on balance, beneficial: it helps reveal unsupported causal jumps, hidden assumptions, and brittle explanations in open-domain reasoning. The same capability, however, can be misused. Counterfactual answers about public policy, historical conflicts, social institutions, disasters, or technological risks may be mistaken for factual predictions, policy advice, or authoritative alternate histories. WhatIfBench is therefore intended only as a diagnostic benchmark, not as a decision-making system. Its questions are collected from publicly accessible sources and normalized for evaluation; we filter out cases that require private or personally identifying information, lack a clear counterfactual premise, or cannot be assessed through an explanatory rubric. Because open-domain sources may still reflect topical, cultural, and English-centric biases, the benchmark should not be treated as exhaustive or culturally neutral. PRISM further relies on automatic parsing and judging of response-derived causal graphs, so its scores measure causal coherence, textual support, and rubric-level explanatory adequacy rather than ground-truth real-world causality. We release WhatIfBench and PRISM for research on counterfactual reasoning evaluation, and recommend that any high-stakes use be paired with human review, inspection of model responses and causal graphs, explicit documentation of source coverage, and safeguards against harmful speculation or unsupported policy conclusions.

\bibliography{custom}

@article{chi2024unveiling,
  title={Unveiling causal reasoning in large language models: Reality or mirage?},
  author={Chi, Haoang and Li, He and Yang, Wenjing and Liu, Feng and Lan, Long and Ren, Xiaoguang and Liu, Tongliang and Han, Bo},
  journal={Advances in Neural Information Processing Systems},
  volume={37},
  pages={96640--96670},
  year={2024}
}

@article{yamin2025can,
  title={Can LLMs Reconcile Knowledge Conflicts in Counterfactual Reasoning},
  author={Yamin, Khurram and Ghosal, Gaurav and Wilder, Bryan},
  journal={arXiv preprint arXiv:2506.15732},
  year={2025}
}

@inproceedings{frohberg2022crass,
  title={CRASS: A novel data set and benchmark to test counterfactual reasoning of large language models},
  author={Frohberg, J{\"o}rg and Binder, Frank},
  booktitle={Proceedings of the Thirteenth Language Resources and Evaluation Conference},
  pages={2126--2140},
  year={2022}
}

@inproceedings{yu2023ifqa,
  title={Ifqa: A dataset for open-domain question answering under counterfactual presuppositions},
  author={Yu, Wenhao and Jiang, Meng and Clark, Peter and Sabharwal, Ashish},
  booktitle={Proceedings of the 2023 Conference on Empirical Methods in Natural Language Processing},
  pages={8276--8288},
  year={2023}
}

@article{chen2025counterbench,
  title={Counterbench: A benchmark for counterfactuals reasoning in large language models},
  author={Chen, Yuefei and Singh, Vivek K and Ma, Jing and Tang, Ruxiang},
  journal={arXiv preprint arXiv:2502.11008},
  year={2025}
}

@article{kiciman2023causal,
  title={Causal reasoning and large language models: Opening a new frontier for causality},
  author={Kiciman, Emre and Ness, Robert and Sharma, Amit and Tan, Chenhao},
  journal={Transactions on Machine Learning Research},
  year={2023}
}

@article{yang2024critical,
  title={A critical review of causal reasoning benchmarks for large language models},
  author={Yang, Linying and Shirvaikar, Vik and Clivio, Oscar and Falck, Fabian},
  journal={arXiv preprint arXiv:2407.08029},
  year={2024}
}

@article{jin2023cladder,
  title={Cladder: Assessing causal reasoning in language models},
  author={Jin, Zhijing and Chen, Yuen and Leeb, Felix and Gresele, Luigi and Kamal, Ojasv and Lyu, Zhiheng and Blin, Kevin and Gonzalez Adauto, Fernando and Kleiman-Weiner, Max and Sachan, Mrinmaya and others},
  journal={Advances in Neural Information Processing Systems},
  volume={36},
  pages={31038--31065},
  year={2023}
}

@inproceedings{wang2024causalbench,
  title={Causalbench: A comprehensive benchmark for evaluating causal reasoning capabilities of large language models},
  author={Wang, Zeyu},
  booktitle={Proceedings of the 10th SIGHAN Workshop on Chinese Language Processing (SIGHAN-10)},
  pages={143--151},
  year={2024}
}

@inproceedings{yu2025causaleval,
  title={Causaleval: Towards better causal reasoning in language models},
  author={Yu, Longxuan and Chen, Delin and Xiong, Siheng and Wu, Qingyang and Li, Dawei and Chen, Zhikai and Liu, Xiaoze and Pan, Liangming},
  booktitle={Proceedings of the 2025 Conference of the Nations of the Americas Chapter of the Association for Computational Linguistics: Human Language Technologies (Volume 1: Long Papers)},
  pages={12512--12540},
  year={2025}
}

@article{lewis1973causation,
  title={Causation},
  author={Lewis, David},
  journal={The journal of philosophy},
  volume={70},
  number={17},
  pages={556--567},
  year={1973},
  publisher={JSTOR}
}

@book{woodward2005making,
  title={Making things happen: A theory of causal explanation},
  author={Woodward, James and Woodward, James Francis},
  year={2005},
  publisher={Oxford university press}
}

@book{tetlock1997counterfactual,
  title={Counterfactual thought experiments in world politics: Logical, methodological, and psychological perspectives},
  author={Tetlock, Philip E and Belkin, Aaron},
  year={1997},
  publisher={Princeton University Press}
}

@misc{glm5team2026glm5vibecodingagentic,
      title={GLM-5: from Vibe Coding to Agentic Engineering},
      author={GLM-5-Team and : and Aohan Zeng and Xin Lv and Zhenyu Hou and Zhengxiao Du and Qinkai Zheng and Bin Chen and Da Yin and Chendi Ge and Chenghua Huang and Chengxing Xie and Chenzheng Zhu and Congfeng Yin and Cunxiang Wang and Gengzheng Pan and Hao Zeng and Haoke Zhang and Haoran Wang and Huilong Chen and Jiajie Zhang and Jian Jiao and Jiaqi Guo and Jingsen Wang and Jingzhao Du and Jinzhu Wu and Kedong Wang and Lei Li and Lin Fan and Lucen Zhong and Mingdao Liu and Mingming Zhao and Pengfan Du and Qian Dong and Rui Lu and Shuang-Li and Shulin Cao and Song Liu and Ting Jiang and Xiaodong Chen and Xiaohan Zhang and Xuancheng Huang and Xuezhen Dong and Yabo Xu and Yao Wei and Yifan An and Yilin Niu and Yitong Zhu and Yuanhao Wen and Yukuo Cen and Yushi Bai and Zhongpei Qiao and Zihan Wang and Zikang Wang and Zilin Zhu and Ziqiang Liu and Zixuan Li and Bojie Wang and Bosi Wen and Can Huang and Changpeng Cai and Chao Yu and Chen Li and Chengwei Hu and Chenhui Zhang and Dan Zhang and Daoyan Lin and Dayong Yang and Di Wang and Ding Ai and Erle Zhu and Fangzhou Yi and Feiyu Chen and Guohong Wen and Hailong Sun and Haisha Zhao and Haiyi Hu and Hanchen Zhang and Hanrui Liu and Hanyu Zhang and Hao Peng and Hao Tai and Haobo Zhang and He Liu and Hongwei Wang and Hongxi Yan and Hongyu Ge and Huan Liu and Huanpeng Chu and Jia'ni Zhao and Jiachen Wang and Jiajing Zhao and Jiamin Ren and Jiapeng Wang and Jiaxin Zhang and Jiayi Gui and Jiayue Zhao and Jijie Li and Jing An and Jing Li and Jingwei Yuan and Jinhua Du and Jinxin Liu and Junkai Zhi and Junwen Duan and Kaiyue Zhou and Kangjian Wei and Ke Wang and Keyun Luo and Laiqiang Zhang and Leigang Sha and Liang Xu and Lindong Wu and Lintao Ding and Lu Chen and Minghao Li and Nianyi Lin and Pan Ta and Qiang Zou and Rongjun Song and Ruiqi Yang and Shangqing Tu and Shangtong Yang and Shaoxiang Wu and Shengyan Zhang and Shijie Li and Shuang Li and Shuyi Fan and Wei Qin and Wei Tian and Weining Zhang and Wenbo Yu and Wenjie Liang and Xiang Kuang and Xiangmeng Cheng and Xiangyang Li and Xiaoquan Yan and Xiaowei Hu and Xiaoying Ling and Xing Fan and Xingye Xia and Xinyuan Zhang and Xinze Zhang and Xirui Pan and Xu Zou and Xunkai Zhang and Yadi Liu and Yandong Wu and Yanfu Li and Yidong Wang and Yifan Zhu and Yijun Tan and Yilin Zhou and Yiming Pan and Ying Zhang and Yinpei Su and Yipeng Geng and Yong Yan and Yonglin Tan and Yuean Bi and Yuhan Shen and Yuhao Yang and Yujiang Li and Yunan Liu and Yunqing Wang and Yuntao Li and Yurong Wu and Yutao Zhang and Yuxi Duan and Yuxuan Zhang and Zezhen Liu and Zhengtao Jiang and Zhenhe Yan and Zheyu Zhang and Zhixiang Wei and Zhuo Chen and Zhuoer Feng and Zijun Yao and Ziwei Chai and Ziyuan Wang and Zuzhou Zhang and Bin Xu and Minlie Huang and Hongning Wang and Juanzi Li and Yuxiao Dong and Jie Tang},
      year={2026},
      eprint={2602.15763},
      archivePrefix={arXiv},
      primaryClass={cs.LG},
      url={https://arxiv.org/abs/2602.15763},
}

@article{yang2025qwen3,
  title={Qwen3 technical report},
  author={Yang, An and Li, Anfeng and Yang, Baosong and Zhang, Beichen and Hui, Binyuan and Zheng, Bo and Yu, Bowen and Gao, Chang and Huang, Chengen and Lv, Chenxu and others},
  journal={arXiv preprint arXiv:2505.09388},
  year={2025}
}

@misc{deepseekai2026deepseekv4,
      title={DeepSeek-V4: Towards Highly Efficient Million-Token Context Intelligence},
      author={DeepSeek-AI},
      year={2026},
}

@techreport{mann1987rhetorical,
  title={Rhetorical structure theory: A theory of text organization},
  author={Mann, William C and Thompson, Sandra A},
  year={1987},
  institution={University of Southern California, Information Sciences Institute Los Angeles}
}

@article{vashishtha2025executable,
  title={Executable Counterfactuals: Improving LLMs' Causal Reasoning Through Code},
  author={Vashishtha, Aniket and Dai, Qirun and Mei, Hongyuan and Sharma, Amit and Tan, Chenhao and Peng, Hao},
  journal={arXiv preprint arXiv:2510.01539},
  year={2025}
}

@article{he2026thinking,
  title={Thinking Fast, Thinking Wrong: Intuitiveness Modulates LLM Counterfactual Reasoning in Policy Evaluation},
  author={He, Yanjie},
  journal={arXiv preprint arXiv:2604.10511},
  year={2026}
}

@inproceedings{bondarenko2022causalqa,
  title={CausalQA: A benchmark for causal question answering},
  author={Bondarenko, Alexander and Wolska, Magdalena and Heindorf, Stefan and Bl{\"u}baum, Lukas and Ngomo, Axel-Cyrille Ngonga and Stein, Benno and Braslavski, Pavel and Hagen, Matthias and Potthast, Martin},
  booktitle={Proceedings of the 29th International Conference on Computational Linguistics},
  pages={3296--3308},
  year={2022}
}

@article{ho2022wikiwhy,
  title={Wikiwhy: Answering and explaining cause-and-effect questions},
  author={Ho, Matthew and Sharma, Aditya and Chang, Justin and Saxon, Michael and Levy, Sharon and Lu, Yujie and Wang, William Yang},
  journal={arXiv preprint arXiv:2210.12152},
  year={2022}
}

@inproceedings{zhang2023causal,
  title={Causal reasoning of entities and events in procedural texts},
  author={Zhang, Li and Xu, Hainiu and Yang, Yue and Zhou, Shuyan and You, Weiqiu and Arora, Manni and Callison-Burch, Chris},
  booktitle={Findings of the Association for Computational Linguistics: EACL 2023},
  pages={415--431},
  year={2023}
}

@article{zheng2025newtonbench,
  title={Newtonbench: Benchmarking generalizable scientific law discovery in llm agents},
  author={Zheng, Tianshi and Tam, Kelvin Kiu-Wai and Nguyen, Newt Hue-Nam K and Xu, Baixuan and Wang, Zhaowei and Cheng, Jiayang and Tsang, Hong Ting and Wang, Weiqi and Bai, Jiaxin and Fang, Tianqing and others},
  journal={arXiv preprint arXiv:2510.07172},
  year={2025}
}

@article{nguyen2025counterfactual,
  title={Counterfactual History: Simulating Chinese Imperial Decisions with AI},
  author={Nguyen, Anh Quang},
  journal={ACTA ILOCANDIA: The DMMMSU International Science and Innovative Technology Journal (Formerly: DMMMSU Research and Extension Journal)},
  volume={9},
  number={1},
  pages={117--138},
  year={2025}
}

@article{wu2024causality,
  title={Causality for large language models},
  author={Wu, Anpeng and Kuang, Kun and Zhu, Minqin and Wang, Yingrong and Zheng, Yujia and Han, Kairong and Li, Baohong and Chen, Guangyi and Wu, Fei and Zhang, Kun},
  journal={arXiv preprint arXiv:2410.15319},
  year={2024}
}

@inproceedings{jin2024can,
  title={Can large language models infer causation from correlation?},
  author={Jin, Zhijing and Liu, Jiarui and Lyu, Zhiheng and Sachan, Mrinmaya and Mihalcea, Rada and Diab, Mona and Ha, David and others},
  booktitle={International Conference on Learning Representations},
  volume={2024},
  pages={28663--28679},
  year={2024}
}

@inproceedings{miliani2025explica,
  title={ExpliCa: Evaluating explicit causal reasoning in large language models},
  author={Miliani, Martina and Auriemma, Serena and Bondielli, Alessandro and Chersoni, Emmanuele and Passaro, Lucia and Sucameli, Irene and Lenci, Alessandro},
  booktitle={Findings of the Association for Computational Linguistics: ACL 2025},
  pages={17335--17355},
  year={2025}
}

@article{yamin2024failure,
  title={Failure modes of llms for causal reasoning on narratives},
  author={Yamin, Khurram and Gupta, Shantanu and Ghosal, Gaurav R and Lipton, Zachary C and Wilder, Bryan},
  journal={arXiv preprint arXiv:2410.23884},
  year={2024}
}

@article{wang2026causalflip,
  title={CausalFlip: A Benchmark for LLM Causal Judgment Beyond Semantic Matching},
  author={Wang, Yuzhe and Zhu, Yaochen and Li, Jundong},
  journal={arXiv preprint arXiv:2602.20094},
  year={2026}
}

@article{balappanawar2025if,
  title={If pigs could fly... can llms logically reason through counterfactuals?},
  author={Balappanawar, Ishwar B and Bonagiri, Vamshi Krishna and Joishy, Anish R and Gaur, Manas and Thirunarayan, Krishnaprasad and Kumaraguru, Ponnurangam},
  journal={arXiv preprint arXiv:2505.22318},
  year={2025}
}

@article{he2026uncovering,
  title={Uncovering Hidden Correctness in LLM Causal Reasoning via Symbolic Verification},
  author={He, Paul and Huang, Yinya and Sachan, Mrinmaya and Jin, Zhijing},
  journal={arXiv preprint arXiv:2601.21210},
  year={2026}
}

@inproceedings{wu2025cofca,
  title={Cofca: A step-wise counterfactual multi-hop qa benchmark},
  author={Wu, Jian and Yang, Linyi and Wang, Zhen and Okumura, Manabu and Zhang, Yue},
  booktitle={International Conference on Learning Representations},
  volume={2025},
  pages={14631--14649},
  year={2025}
}

@inproceedings{du2025confidence,
  title={Confidence Calibration for Multimodal LLMs: An Empirical Study through Medical VQA},
  author={Du, Yuetian and Wang, Yucheng and Kong, Ming and Liang, Tian and Long, Qiang and Chen, Bingdi and Zhu, Qiang},
  booktitle={International Conference on Medical Image Computing and Computer-Assisted Intervention},
  pages={89--99},
  year={2025},
  organization={Springer}
}

@article{du2026reveal,
  title={REVEAL: Reverse Engineering Agentic Latents from Frontier-Model Traces},
  author={Du, Yuetian and Wang, Meng and Wang, Yucheng and Chen, Zhangzhe and Wang, Wei and Yang, Boyu and Zhai, Weiqi and Zhao, Bing and Kong, Ming and Zhou, Zhan and others},
  journal={arXiv preprint arXiv:2610.16288},
  year={2026}
}

@InProceedings{Du_2026_CVPR,
    author    = {Du, Yuetian and Wang, Yucheng and Zhang, Rongyu and Xu, Zhijie and Yang, Boyu and Kong, Ming and Liu, Jie and Zhu, Qiang},
    title     = {Linking Perception, Confidence and Accuracy in MLLMs},
    booktitle = {Proceedings of the IEEE/CVF Conference on Computer Vision and Pattern Recognition (CVPR)},
    month     = {June},
    year      = {2026},
    pages     = {25914-25924}
}

@inproceedings{du2026care,
  title={CARE: Confidence-Aware Reasoning for Reliable Medical VQA},
  author={Du, Yuetian and Wang, Yucheng and Chen, Zhenyuan and Chen, Luyuan and Zhang, Rongyu and Zhang, Jinjian and Zhou, Wei and Xu, Zhijie and Kong, Ming and Zhou, Zhan and others},
  booktitle={International Conference on Medical Image Computing and Computer-Assisted Intervention},
  pages={66--75},
  year={2026},
  organization={Springer}
}

\newpage
\clearpage

\appendix

\section*{Appendix}

In this appendix, we provide supplementary materials for \textbf{WhatIfBench} and \textbf{PRISM} to improve the transparency and reproducibility of our benchmark and evaluation framework. We first present additional dataset statistics in Appendix~\ref{app:dataset_stats}, including the overall scenario taxonomy, fine-grained category distributions, and lexical characteristics of WhatIfBench. We then describe the data collection, annotation, and quality-control process in Appendix~\ref{app:data_annotation}, covering candidate sources, filtering criteria, rubric construction guidelines, and review protocols. Appendices~\ref{app:decoding_settings} and~\ref{app:prompts} provides the complete prompt templates used for model response generation, response-to-graph parsing, Process Metric evaluation, and Rubric Metric evaluation, together with decoding and evaluator settings. Appendix~\ref{app:parsing_algorithm} further details the response-to-graph parsing procedure and presents the parser pseudocode. Appendix~\ref{app:additional_experiments} reports additional experiments on benchmark positioning, model scale, response length, paraphrase robustness, prompting strategies, and evaluator-family bias. Finally, Appendix~\ref{app:worked_example} gives a complete worked example of PRISM evaluation, including a what-if question with its query-dependent rubrics, a model response, the derived semantic causal graph, and RM judgment.

\section{Dataset Details and Statistics}
\label{app:dataset_stats}

\subsection{Scenario Taxonomy}

\begin{figure}[t]
    \centering
    \includegraphics[width=0.58\linewidth]{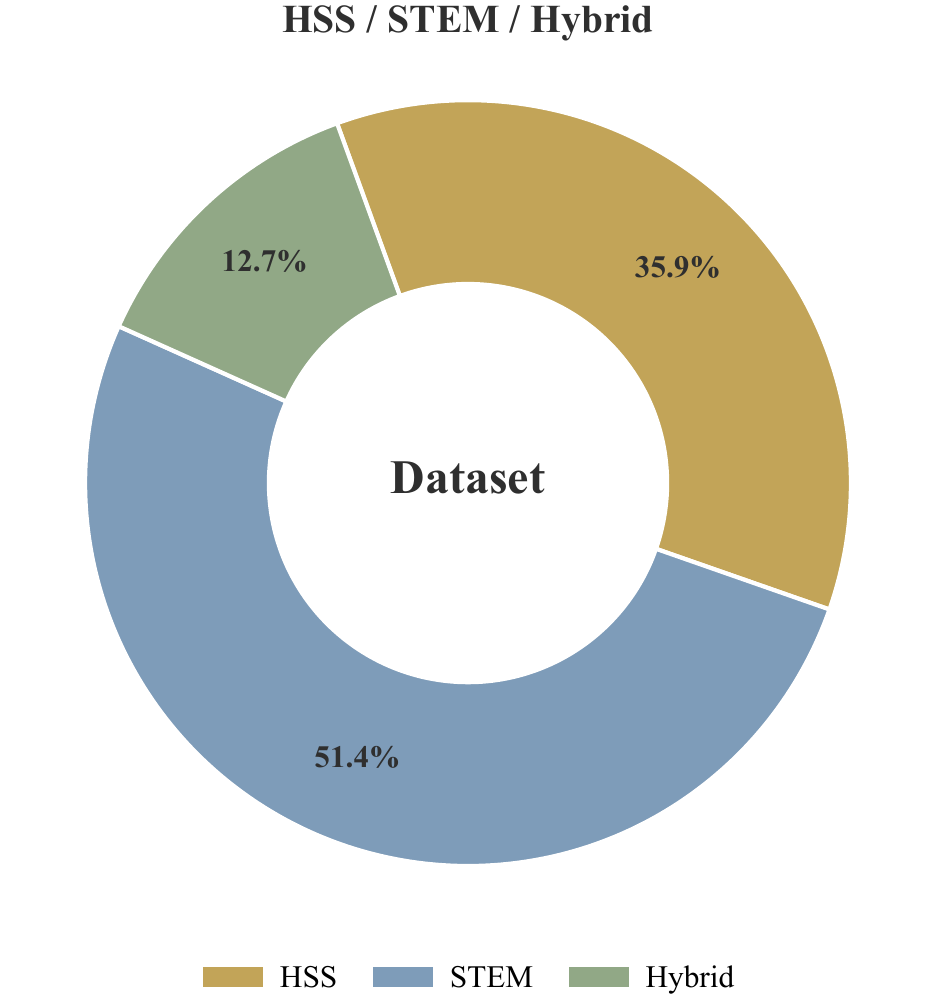}
    \caption{\textbf{Scenario distribution of WhatIfBench.} The benchmark contains 220 open-form what-if questions across STEM, HSS, and Hybrid scenarios. STEM accounts for 51.4\% of the dataset, followed by HSS at 35.9\% and Hybrid at 12.7\%.}
    \label{fig:app_scenario_taxonomy}
\end{figure}

WhatIfBench organizes open-form counterfactual questions into three scenario types: STEM, HSS, and Hybrid. STEM scenarios focus on scientific, technical, physical, biological, and engineering mechanisms, where counterfactual consequences are often constrained by relatively stable domain principles. HSS scenarios cover historical, political, social, cultural, and institutional dynamics, where reasoning must account for path dependence, multi-agent behavior, and long-term social feedback. Hybrid scenarios involve cross-domain propagation, where changes in natural, technical, or material conditions further affect social, economic, or institutional systems. As shown in Figure~\ref{fig:app_scenario_taxonomy}, STEM constitutes the largest portion of WhatIfBench, followed by HSS and Hybrid, reflecting our emphasis on both mechanistic reasoning and socio-historical causal propagation.

\subsection{Fine-grained Category Distribution}

We further group STEM and HSS questions into fine-grained subcategories to characterize the topical coverage of WhatIfBench. For HSS, the largest group is State, Regime, and Macrohistory, followed by War and Geopolitics, while the remaining questions cover dynastic leadership, domestic politics and institutions, and culture, technology, and society. For STEM, Physical and Planetary Systems is the largest group, followed by Engineering, Transport, and Design, Society, Computation, and Logistics, and Life Sciences and Human Body. These subcategories are not intended as exhaustive domain taxonomies; rather, they provide diagnostic slices for analyzing whether models fail more often on physical mechanisms, biological constraints, institutional dynamics, geopolitical reasoning, or cross-system dependencies.

\begin{figure}[t]
    \centering
    \includegraphics[width=0.58\linewidth]{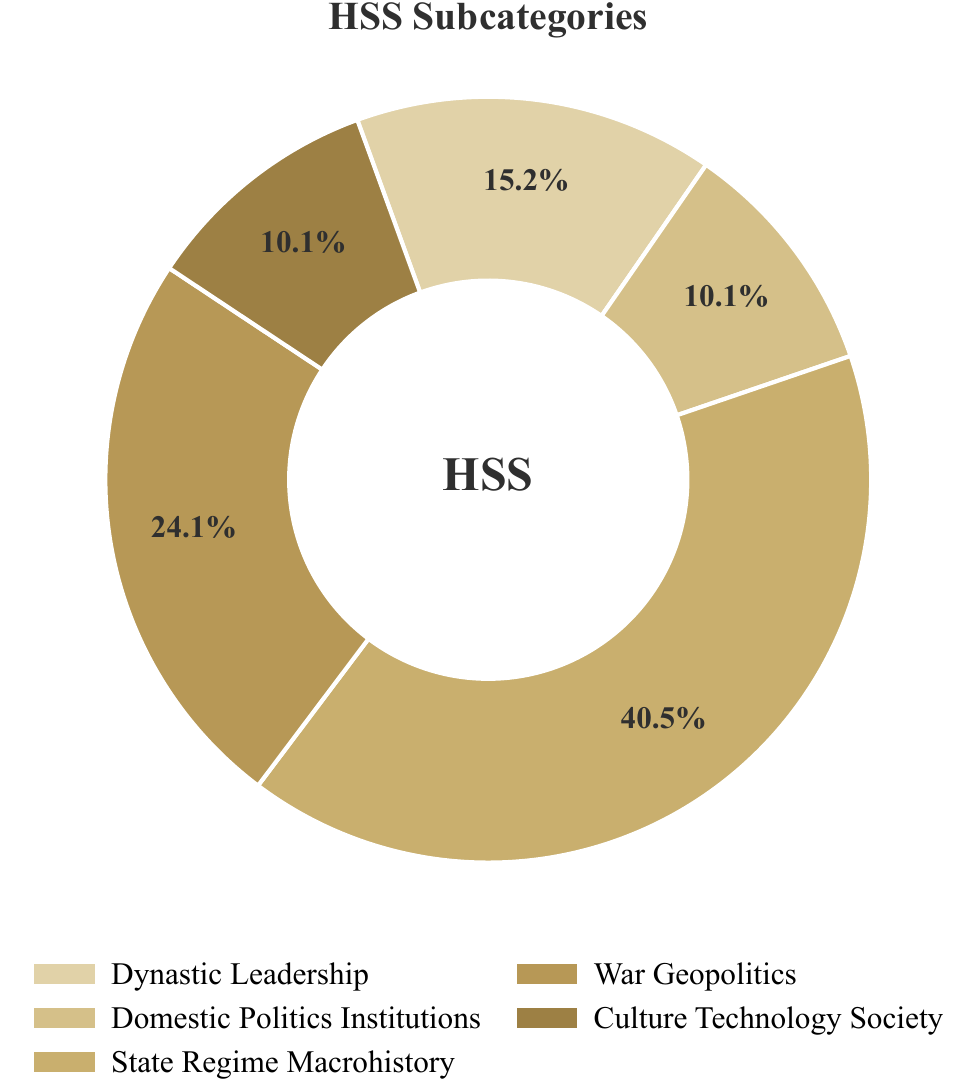}
    \caption{\textbf{Fine-grained distribution of HSS questions.} HSS scenarios are divided into State, Regime, and Macrohistory; War and Geopolitics; Dynastic Leadership; Domestic Politics and Institutions; and Culture, Technology, and Society. State, Regime, and Macrohistory is the largest HSS subcategory at 40.5\%.}
    \label{fig:app_hss_subcategories}
\end{figure}

\begin{figure}[t]
    \centering
    \includegraphics[width=0.58\linewidth]{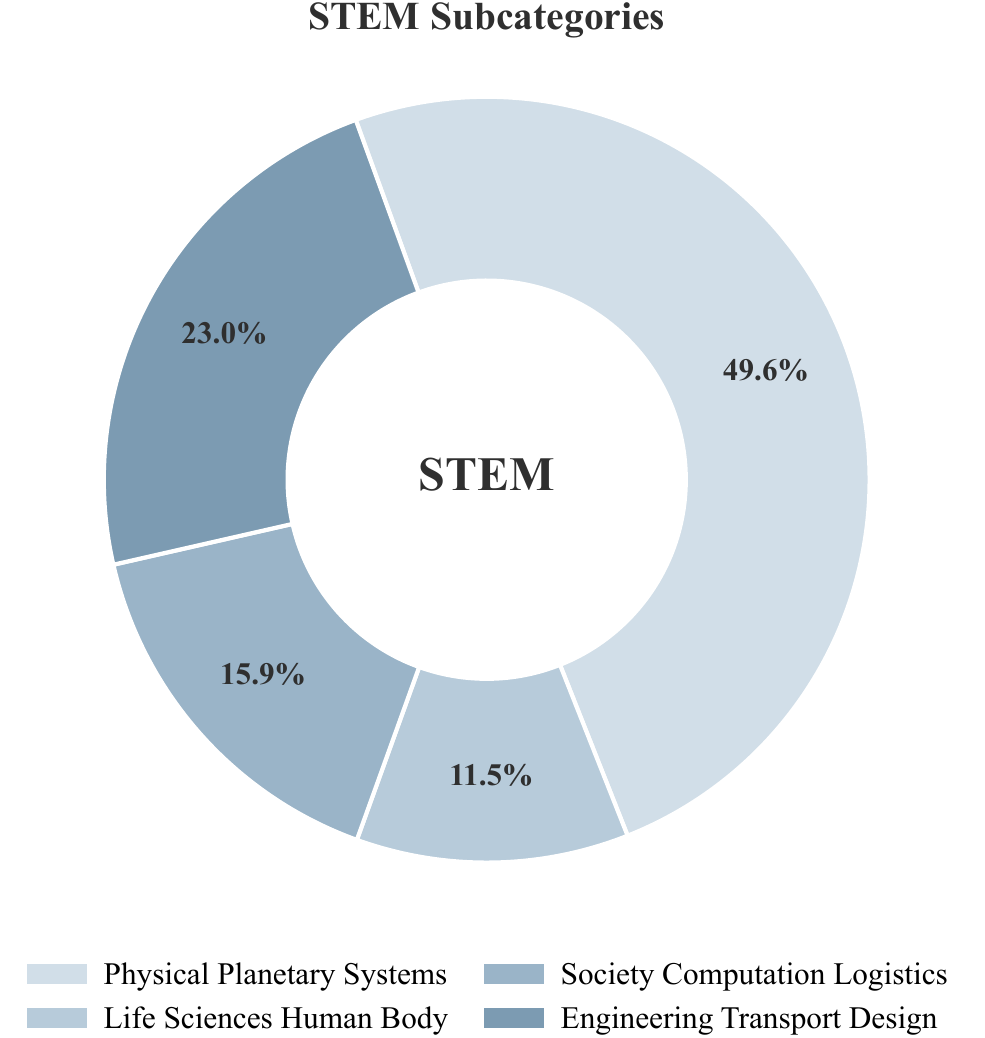}
    \caption{\textbf{Fine-grained distribution of STEM questions.} STEM scenarios are divided into Physical and Planetary Systems, Life Sciences and Human Body, Society, Computation, and Logistics, and Engineering, Transport, and Design. Physical and Planetary Systems forms the largest STEM subcategory at 49.6\%.}
    \label{fig:app_stem_subcategories}
\end{figure}

\subsection{Lexical Overview}

\begin{figure}[t]
    \centering
    \includegraphics[width=0.9\linewidth]{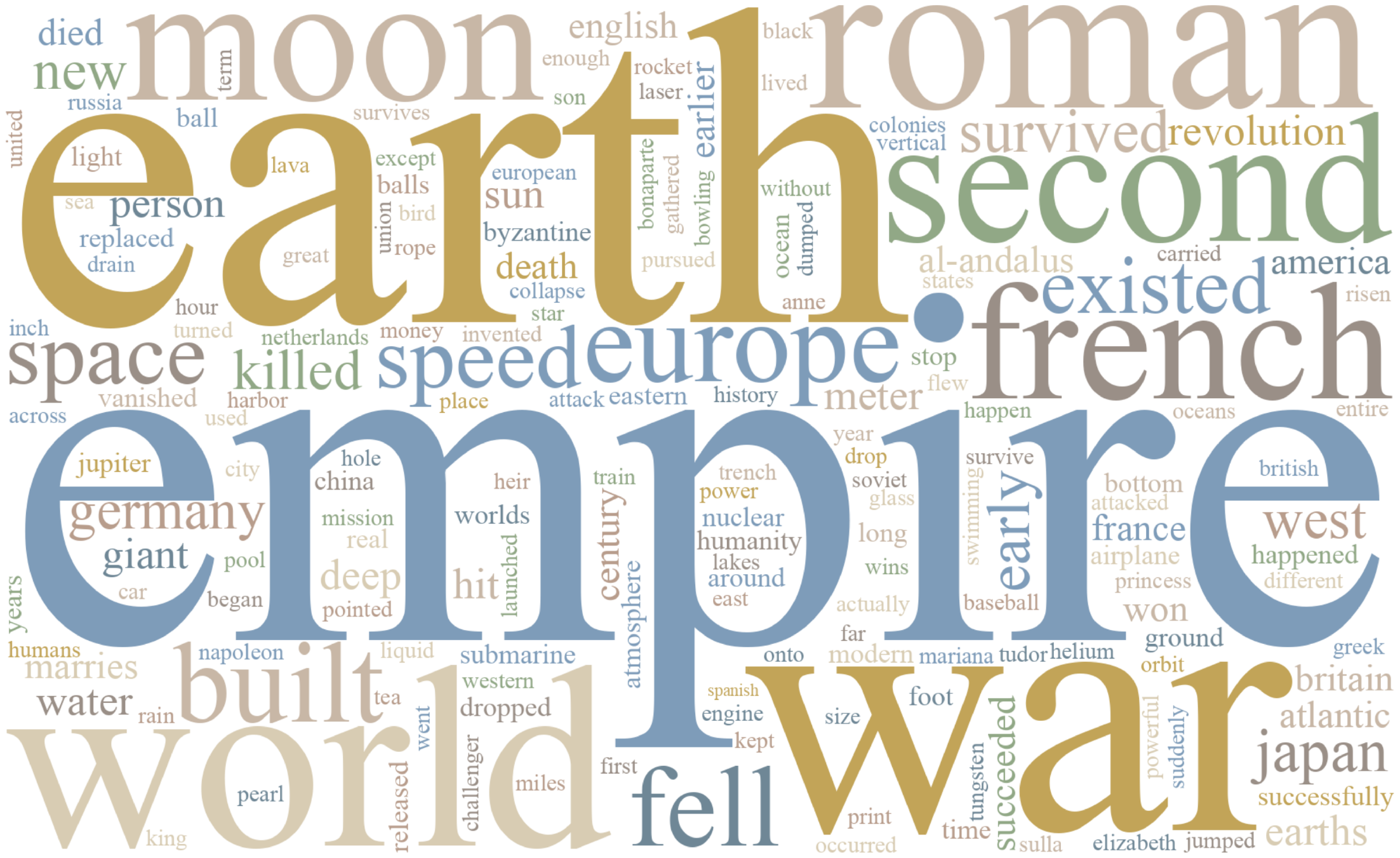}
    \caption{\textbf{Lexical overview of WhatIfBench questions.} The word cloud visualizes frequent terms in normalized what-if questions, offering a qualitative view of the benchmark's topical coverage across scientific, historical, geopolitical, technological, and cross-domain scenarios.}
    \label{fig:app_wordcloud}
\end{figure}

To provide an intuitive view of the topical coverage of WhatIfBench, we visualize the lexical distribution of normalized questions in Figure~\ref{fig:app_wordcloud}. The most frequent terms reflect the benchmark's mixture of scientific, historical, geopolitical, and technological scenarios, including planetary systems, empires, wars, societies, and alternative historical trajectories. This lexical overview is intended as a qualitative summary rather than a formal diversity measure; the structured scenario taxonomy and fine-grained category distributions provide the primary basis for dataset characterization.

\section{Data Collection, Annotation, and Quality Control}
\label{app:data_annotation}

\subsection{Candidate Collection}

We construct WhatIfBench from publicly accessible what-if questions and explanatory discussions. Candidate questions are collected from public what-if QA sites, online discussion forums, and explanatory materials, including xkcd What If, Worldbuilding Stack Exchange, AlternateHistory, and Quora. These sources contain naturally occurring counterfactual questions that often require open-form explanations rather than short factual answers. During collection, we retain the original counterfactual intent of each question while normalizing its wording into a clear query format suitable for model evaluation.

\subsection{Question Filtering Criteria}

We filter candidate questions according to four criteria. First, each question must contain an explicit counterfactual premise or intervention. Second, the question should require nontrivial causal propagation rather than simple factual lookup or direct answer retrieval. Third, the question should admit open-form explanation, allowing multiple plausible downstream trajectories rather than a single fixed outcome. Fourth, the question should provide sufficient contextual grounding for constructing a meaningful rubric and evaluating the response. We reject questions that lack a clear counterfactual premise, depend primarily on private or personally identifying information, are too underspecified for rubric-based evaluation, or cannot be assessed through explanatory criteria.

\subsection{Rubric Construction Guidelines}

For each retained question, we construct a query-specific rubric to specify the reasoning requirements expected of a high-quality counterfactual explanation. Each rubric is designed to cover the key dimensions used by PRISM, including fidelity to the counterfactual premise, coverage of central mechanisms, treatment of downstream consequences, respect for domain constraints, uncertainty handling, and overall explanatory completeness. The rubric does not prescribe a unique conclusion. Instead, it defines question-dependent requirements that different plausible answers should satisfy, allowing evaluation to focus on whether the response supports its trajectory with coherent causal reasoning.

\begin{table*}[t]
\centering
\small
\setlength{\tabcolsep}{7pt}
\renewcommand{\arraystretch}{1.1}
\begin{tabular}{lccccc}
\toprule
\textbf{Benchmark} &
\shortstack{\textbf{Open-}\\\textbf{domain}} &
\shortstack{\textbf{Open-}\\\textbf{form}} &
\shortstack{\textbf{Long-}\\\textbf{horizon}} &
\shortstack{\textbf{Process-level}\\\textbf{Evaluation}} &
\shortstack{\textbf{Question-specific}\\\textbf{Rubrics}} \\
\midrule
CRASS~\citep{frohberg2022crass}
& $\times$ & $\times$ & $\times$ & $\times$ & $\times$ \\
IfQA~\citep{yu2023ifqa}
& \checkmark & $\times$ & $\times$ & $\times$ & $\times$ \\
CLadder~\citep{jin2023cladder}
& $\times$ & $\times$ & $\times$ & $\times$ & $\times$ \\
CofCA~\citep{wu2025cofca}
& $\times$ & $\times$ & $\times$ & \checkmark & $\times$ \\
CounterBench~\citep{chen2025counterbench}
& $\times$ & $\times$ & \checkmark & $\times$ & $\times$ \\
\textbf{WhatIfBench}
& \checkmark & \checkmark & \checkmark & \checkmark & \checkmark \\
\bottomrule
\end{tabular}
\caption{\textbf{Comparison of counterfactual reasoning benchmark characteristics.}}
\label{tab:benchmark_characteristics}
\end{table*}

\begin{table*}[t]
\centering
\small
\setlength{\tabcolsep}{7pt}
\renewcommand{\arraystretch}{1.08}
\begin{tabular}{lcccccc}
\toprule
\multirow{2}{*}{\textbf{Model}} &
\multicolumn{2}{c}{\textbf{WhatIfBench}} &
\multicolumn{2}{c}{\textbf{IfQA}} &
\multicolumn{2}{c}{\textbf{CounterBench}} \\
\cmidrule(lr){2-3}
\cmidrule(lr){4-5}
\cmidrule(lr){6-7}
& \textbf{Final} $\uparrow$ & \textbf{Rank}
& \textbf{EM} $\uparrow$ & \textbf{Rank}
& \textbf{Acc.} $\uparrow$ & \textbf{Rank} \\
\midrule
GLM-5.1                & 51.99 & 6 & 89.71 & 6 & 80.00 & 3 \\
DeepSeek-V4-Pro        & 55.12 & 4 & 90.43 & 4 & 85.71 & 1 \\
Qwen3-Max              & 52.26 & 5 & 93.00 & 1 & 76.74 & 6 \\
Gemini-3.1-Pro-Preview & 56.33 & 3 & 90.00 & 5 & 78.33 & 5 \\
Claude-Opus-4.7        & 59.11 & 2 & 91.86 & 3 & 79.97 & 4 \\
GPT-5.5                & 64.62 & 1 & 92.57 & 2 & 83.57 & 2 \\
\bottomrule
\end{tabular}
\caption{\textbf{Performance and rankings of frontier LLMs across counterfactual reasoning benchmarks.}}
\label{tab:cross_benchmark_results}
\end{table*}

\subsection{Quality Control Protocol}

We conduct quality control at both the question and rubric levels. For questions, review checks whether the normalized query has a clear counterfactual premise, sufficient contextual grounding, nontrivial causal propagation requirements, and an evaluable explanatory structure. For rubrics, review checks whether the criteria cover the core reasoning space without forcing a single gold trajectory. Each item is assigned one of three outcomes: \textsc{Pass}, if it can be used directly; \textsc{Revised}, if it becomes valid after normalization or rubric refinement; and \textsc{Rejected}, if it does not meet the benchmark requirements. As reported in the main text, 36.0\% of 611 candidate questions are retained after direct pass or revision, yielding 220 final questions, and all corresponding rubric sets pass after direct review or revision.

\section{Decoding and Evaluator Settings}
\label{app:decoding_settings}

To ensure consistent text generation and rigorous evaluation within the \textbf{PRISM} framework, we establish a standardized set of decoding hyper-parameters and structural parsing rules across our entire pipeline. 

For the initial \textit{Model Response Generation} phase on \textbf{WhatIfBench}, models generate counterfactual answers using a sampling temperature of $\tau = 0.6$ and a maximum budget of $4,096$ tokens to encourage natural, comprehensive, and descriptive long-range causal explanations. Conversely, for all downstream parsing and judgment tasks, we enforce a strictly deterministic decoding policy by setting the temperature to $\tau = 0.0$ across all evaluation API calls. The maximum generation budget varies depending on the structural density required by each core component: the \textit{Response-to-Graph Parser} is allocated up to $16,384$ tokens to accommodate construction of a full \textit{Response-Derived Semantic Causal Graph}, whereas the \textit{Process Metric (PM) Edge Evaluator} and \textit{Rubric Metric (RM) Evaluator} are capped at $1,024$ and $4,096$ tokens, respectively.

\section{Prompt Templates}
\label{app:prompts}

For complete reproducibility, we detail the prompt templates (\cref{fig:model_response_prompt,fig:response_to_graph_parser_prompt,fig:pm_edge_evaluation_prompt,fig:rubric_evaluation_prompt}) used for model response generation, graph parsing, and dual-metric evaluation within the \textbf{PRISM} framework.

\section{Response-to-Graph Parsing Algorithm}
\label{app:parsing_algorithm}

Algorithm~\ref{alg:response_to_graph} summarizes how we convert each model response $y$ into a response-derived semantic causal graph $\mathcal{G}_{y}=(\mathcal{V}_{y},\mathcal{E}_{y})$. The parser first extracts elementary discourse units (EDUs) and semantic relation records with labels from the predefined set $\mathcal{L}$. These records are normalized and organized into a compact support--core structure. We then project the structure onto EDU-level graph nodes, where each resolved relation becomes a directed edge after duplicate removal. If parsing fails or no valid EDU is extracted, we use a conservative heuristic fallback that splits the response into sentence-level EDUs and connects adjacent units with $\mathrm{SEQUENCE}$ edges. This keeps the graph construction grounded in the response text while ensuring that all responses can be evaluated under a consistent graph format.

\begin{algorithm}[t]
\small
\caption{Response-to-Graph Parsing}
\label{alg:response_to_graph}
\begin{algorithmic}[1]
\Require Model response $y$, parser model $p_{\theta}$, relation labels $\mathcal{L}$, budget $B$
\Ensure Response-Derived Semantic Causal Graph $\mathcal{G}_{y}=(\mathcal{V}_{y},\mathcal{E}_{y})$

\State Construct prompt $m(y,\mathcal{L})$ for extracting EDUs, semantic relations, and support--core structure.
\State Obtain parser output $o \leftarrow p_{\theta}(m(y,\mathcal{L}); \tau=0, B)$.
\If{$o$ is empty}
    \State \Return $\textsc{HeuristicGraph}(y)$
\EndIf

\State Parse EDUs $\mathcal{U}=\{u_i\}_{i=1}^{n}$ from $o$.
\If{$\mathcal{U}=\emptyset$}
    \State \Return $\textsc{HeuristicGraph}(y)$
\EndIf

\State Parse semantic relations $\mathcal{R}=\{(s_j,t_j,\ell_j,\nu_j)\}_{j=1}^{m}$, where $\ell_j\in\mathcal{L}$ and $\nu_j\in\{\mathrm{SN},\mathrm{NS},\mathrm{NN}\}$.
\If{$\mathcal{R}\neq\emptyset$}
    \State Build structure nodes $\mathcal{T}\leftarrow\textsc{BuildStructure}(\mathcal{R})$.
\ElsIf{$|\mathcal{U}|>1$}
    \State Build a flat fallback structure $\mathcal{T}\leftarrow\textsc{BuildFlat}(\mathcal{U},\mathrm{SEQUENCE})$.
\Else
    \State $\mathcal{T}\leftarrow\emptyset$.
\EndIf

\State Initialize graph nodes $\mathcal{V}_{y}\leftarrow\{(u_i,\mathrm{text}(u_i))\mid u_i\in\mathcal{U}\}$ and edges $\mathcal{E}_{y}\leftarrow\emptyset$.
\State Build reference map $\mathcal{M}$ from EDU spans and structure nodes to their resolved EDUs.

\ForAll{structure node $z\in\mathcal{T}$}
    \State Normalize relation label $\ell\leftarrow\textsc{Normalize}(z.\mathrm{relation})$.
    \State Resolve source EDUs $\mathcal{S}$ and target EDUs $\mathcal{D}$ from $z$ using $\mathcal{M}$.
    \ForAll{$u_s\in\mathcal{S}, u_t\in\mathcal{D}$}
        \If{$u_s\neq u_t$ and $(u_s,u_t,\ell)$ is not duplicated}
            \State Add edge $(u_s,u_t,\ell)$ to $\mathcal{E}_{y}$.
        \EndIf
    \EndFor
\EndFor

\State \Return $\mathcal{G}_{y}=(\mathcal{V}_{y},\mathcal{E}_{y})$

\vspace{1mm}
\Function{HeuristicGraph}{$y$}
    \State Split $y$ into sentence-level EDUs and connect adjacent EDUs with $\mathrm{SEQUENCE}$ edges.
    \State \Return the resulting graph.
\EndFunction
\end{algorithmic}
\end{algorithm}

\section{Additional Experiments}
\label{app:additional_experiments}

We conduct additional experiments to examine the positioning and robustness of WhatIfBench and PRISM, including comparisons with existing benchmarks, smaller-model evaluation, response-length control, paraphrase stress testing, counterfactual-specific prompting, and cross-family evaluation.

\noindent\textbf{Comparison with existing counterfactual benchmarks.}
Table~\ref{tab:benchmark_characteristics} compares WhatIfBench with representative counterfactual reasoning benchmarks. Existing benchmarks mainly use constrained answer spaces or fixed outcomes, whereas WhatIfBench jointly targets open-domain, open-form, long-horizon reasoning with process-level evaluation and question-specific rubrics.

We further evaluate the same six frontier models on IfQA and CounterBench. As shown in Table~\ref{tab:cross_benchmark_results}, performance is substantially higher and more compressed on the closed-form benchmarks, while model rankings differ considerably across tasks.

\begin{table*}[t]
\centering
\small
\resizebox{\textwidth}{!}{%
\begin{tabular}{llccccccccc}
\toprule
\textbf{Model} &
\textbf{Architecture} &
\multicolumn{2}{c}{\textbf{STEM}} &
\multicolumn{2}{c}{\textbf{HSS}} &
\multicolumn{2}{c}{\textbf{Hybrid}} &
\multicolumn{2}{c}{\textbf{Overall}} &
\textbf{Final} \\
\cmidrule(lr){3-4}
\cmidrule(lr){5-6}
\cmidrule(lr){7-8}
\cmidrule(lr){9-10}
& & PM & RM & PM & RM & PM & RM & PM & RM & \\
\midrule
Qwen3-4B       & Dense & 48.87 & 36.18 & 45.72 & 32.88 & 39.16 & 36.80 & 46.59 & 35.02 & 40.80 \\
Qwen3-8B       & Dense & 53.08 & 34.48 & 42.49 & 32.54 & 55.10 & 37.20 & 49.36 & 34.06 & 41.71 \\
Gemma-3-12B-it & Dense & 49.45 & 41.73 & 43.92 & 41.61 & 44.72 & 43.28 & 46.85 & 41.86 & 44.36 \\
GPT-oss-20B    & MoE   & 52.47 & 43.37 & 42.76 & 36.54 & 46.26 & 38.64 & 48.20 & 40.41 & 44.26 \\
\bottomrule
\end{tabular}%
}
\caption{\textbf{Performance of smaller models on WhatIfBench across model architectures.}}
\label{tab:smaller_models}
\end{table*}

Qwen3-Max ranks first on IfQA but fifth on WhatIfBench, while DeepSeek-V4-Pro ranks first on CounterBench but fourth on WhatIfBench. As an external comparison, WhatIfBench also shows stronger rank agreement with the Artificial Analysis Intelligence Index\footnote{\url{https://artificialanalysis.ai/evaluations/artificial-analysis-intelligence-index}} ($\rho=0.94$) than IfQA ($\rho=0.09$) or CounterBench ($\rho=0.37$), suggesting that these benchmarks capture complementary capabilities.

\noindent\textbf{Evaluation on smaller models.}
To examine discriminability beyond frontier models, we evaluate four smaller models spanning dense and MoE architectures. Their final scores range from 40.80 to 44.36, below the frontier-model range of 51.99--64.62, showing that WhatIfBench remains discriminative across model scales and architectures.

\noindent\textbf{Effect of response length.}
PRISM does not explicitly reward response length, but longer outputs may introduce more causal content. We therefore evaluate GPT-5.5 under Vanilla, Long Output, and Short Output settings. The final-score standard deviation is only 0.48, indicating limited sensitivity to moderate changes in response length.

\begin{table*}[t]
\centering
\small
\resizebox{0.8\textwidth}{!}{%
\begin{tabular}{lccccccccc}
\toprule
\textbf{Method} &
\multicolumn{2}{c}{\textbf{STEM}} &
\multicolumn{2}{c}{\textbf{HSS}} &
\multicolumn{2}{c}{\textbf{Hybrid}} &
\multicolumn{2}{c}{\textbf{Overall}} &
\textbf{Final} \\
\cmidrule(lr){2-3}
\cmidrule(lr){4-5}
\cmidrule(lr){6-7}
\cmidrule(lr){8-9}
& PM & RM & PM & RM & PM & RM & PM & RM & \\
\midrule
Vanilla
& 72.77 & 64.00 & 57.38 & 61.98 & 65.32 & 64.80 & 65.94 & 63.31 & 64.62 \\
Long Output
& 73.17 & 64.42 & 57.13 & 61.87 & 67.49 & 65.60 & 66.69 & 63.73 & 65.21 \\
Short Output
& 73.31 & 64.30 & 58.35 & 62.46 & 69.65 & 65.20 & 67.47 & 63.69 & 65.58 \\
\midrule
SD
& 0.28 & 0.22 & 0.64 & 0.31 & 2.17 & 0.40 & 0.77 & 0.23 & 0.48 \\
\bottomrule
\end{tabular}%
}
\caption{\textbf{Effect of response length on PRISM scores.}
SD is computed across the three response-length settings.}
\label{tab:length_control}
\end{table*}

\begin{table*}[t]
\centering
\small
\resizebox{0.8\textwidth}{!}{%
\begin{tabular}{lccccccccc}
\toprule
\textbf{Method} &
\multicolumn{2}{c}{\textbf{STEM}} &
\multicolumn{2}{c}{\textbf{HSS}} &
\multicolumn{2}{c}{\textbf{Hybrid}} &
\multicolumn{2}{c}{\textbf{Overall}} &
\textbf{Final} \\
\cmidrule(lr){2-3}
\cmidrule(lr){4-5}
\cmidrule(lr){6-7}
\cmidrule(lr){8-9}
& PM & RM & PM & RM & PM & RM & PM & RM & \\
\midrule
Paraphrase
& 2.65 & 10.00 & 1.83 & 10.10 & 0.00 & 10.00 & 2.05 & 10.04 & 6.04 \\
\bottomrule
\end{tabular}%
}
\caption{\textbf{PRISM stress test with paraphrase-only responses.}}
\label{tab:paraphrase_stress}
\end{table*}

\noindent\textbf{Paraphrase-only stress test.}
We test whether simply restating the counterfactual question can receive substantial PRISM credit. Paraphrase-only responses obtain 2.05 PM and 6.04 Final. The RM value near 10 corresponds to the minimum score of 1 on the 1--10 rubric scale, confirming that surface reformulation alone receives negligible credit.

\noindent\textbf{Counterfactual-specific prompting.}
The main experiments use a standardized prompt for controlled comparison. We additionally evaluate GPT-5.5 with two counterfactual-specific strategies, CausalCoT and CoIn. CausalCoT and CoIn improve the final score by 2.63 and 1.35 points, respectively. The gains are moderate and affect PM and RM differently, indicating that prompting helps but does not eliminate the remaining difficulty. Under the prompting strategies evaluated here, WhatIfBench remains challenging for frontier models.

\begin{table*}[t]
\centering
\small
\resizebox{0.8\textwidth}{!}{%
\begin{tabular}{lccccccccc}
\toprule
\textbf{Method} &
\multicolumn{2}{c}{\textbf{STEM}} &
\multicolumn{2}{c}{\textbf{HSS}} &
\multicolumn{2}{c}{\textbf{Hybrid}} &
\multicolumn{2}{c}{\textbf{Overall}} &
\textbf{Final} \\
\cmidrule(lr){2-3}
\cmidrule(lr){4-5}
\cmidrule(lr){6-7}
\cmidrule(lr){8-9}
& PM & RM & PM & RM & PM & RM & PM & RM & \\
\midrule
Vanilla
& 72.77 & 64.00 & 57.38 & 61.98 & 65.32 & 64.80 & 65.94 & 63.31 & 64.62 \\
CausalCoT
& 63.98 & 68.34 & 68.32 & 67.88 & 68.78 & 70.00 & 66.12 & 68.37 & 67.25 \\
CoIn
& 69.16 & 62.63 & 69.60 & 61.15 & 73.65 & 64.84 & 69.61 & 62.33 & 65.97 \\
\bottomrule
\end{tabular}%
}
\caption{\textbf{Effect of counterfactual-specific prompting strategies on GPT-5.5.}}
\label{tab:prompting_strategies}
\end{table*}

\noindent\textbf{Robustness to evaluator family.}
To test potential same-family evaluator bias, we re-evaluate all six frontier models using Claude-Sonnet-5 under the same PRISM protocol. The two evaluators produce identical model rankings (Spearman $\rho=1.00$). The mean absolute difference in final score is 0.63 points, with a maximum of 1.57 points, suggesting that the main comparative conclusions are robust to the evaluator family.

\begin{table*}[t]
\centering
\small
\resizebox{\textwidth}{!}{%
\begin{tabular}{lcccccccccc}
\toprule
\textbf{Model} &
\multicolumn{2}{c}{\textbf{STEM}} &
\multicolumn{2}{c}{\textbf{HSS}} &
\multicolumn{2}{c}{\textbf{Hybrid}} &
\multicolumn{2}{c}{\textbf{Overall}} &
\textbf{Final} &
\textbf{Rank} \\
\cmidrule(lr){2-3}
\cmidrule(lr){4-5}
\cmidrule(lr){6-7}
\cmidrule(lr){8-9}
& PM & RM & PM & RM & PM & RM & PM & RM & & \\
\midrule
GLM-5.1
& 59.59 & 52.19 & 48.64 & 47.00 & 53.95 & 50.32 & 54.81 & 49.99 & 52.40 & 6 \\
DeepSeek-V4-Pro
& 65.75 & 55.50 & 50.91 & 52.94 & 42.05 & 54.75 & 57.51 & 54.42 & 55.97 & 4 \\
Qwen3-Max
& 67.75 & 48.27 & 53.99 & 43.00 & 56.85 & 48.56 & 61.33 & 46.34 & 53.83 & 5 \\
Gemini-3.1-Pro-Preview
& 67.89 & 54.18 & 51.57 & 50.32 & 50.89 & 50.24 & 59.83 & 52.29 & 56.06 & 3 \\
Claude-Opus-4.7
& 68.88 & 56.11 & 56.89 & 52.37 & 55.57 & 58.88 & 62.87 & 54.94 & 58.90 & 2 \\
GPT-5.5
& 73.51 & 64.10 & 57.85 & 62.02 & 63.20 & 65.68 & 66.57 & 63.55 & 65.06 & 1 \\
\bottomrule
\end{tabular}%
}
\caption{\textbf{PRISM scores and model rankings using Claude-Sonnet-5 as the evaluator.}}
\label{tab:cross_family_judge}
\end{table*}

\section{Worked Example of PRISM Evaluation}
\label{app:worked_example}

For case-level transparency, we provide a worked example of \textbf{PRISM} evaluation on one WhatIfBench question. \Cref{fig:case_question_rubric} presents the counterfactual question and its query-dependent rubric, while \cref{fig:case_model_response} shows the generated model response. PRISM then parses the response into a response-derived semantic causal graph, visualized in \cref{fig:case_response_graph_visual}, where related EDUs are grouped into causal modules for readability. \Cref{fig:case_rubric_judgment} reports the RM judgment with criterion-level scores and rationales. The example shows how PRISM jointly exposes the response's expressed reasoning structure and its answer-level omissions: although the response contains plausible local causal links, it receives a moderate RM score because the point of divergence, institutional integration pathway, and uncertainty control remain underdeveloped.

\begin{figure*}[t]
\centering
\begin{tcolorbox}[
    width=0.9\textwidth,
    colback=gray!2,
    colframe=gray!45,
    boxrule=0.5pt,
    arc=2pt,
    left=6pt,
    right=6pt,
    top=6pt,
    bottom=6pt,
    title={Model Response Generation Prompt},
    fonttitle=\bfseries,
    coltitle=black,
    colbacktitle=gray!12
]
\small
\noindent\textbf{System Prompt:}

\vspace{0.3em}
\noindent
You answer counterfactual benchmark questions. Give a direct, coherent, non-bulleted answer unless bullets are clearly helpful.

\vspace{0.8em}
\noindent\textbf{User Prompt:}

\vspace{0.3em}
\noindent
Question domain: \{domain\}\\
Question type: \{answer\_type\}\\[0.6em]
Question:\\
\{question\}\\[0.6em]
Answer the question carefully and directly.
\end{tcolorbox}
\caption{Model response generation prompt used for WhatIfBench evaluation.}
\label{fig:model_response_prompt}
\end{figure*}

\begin{figure*}[t]
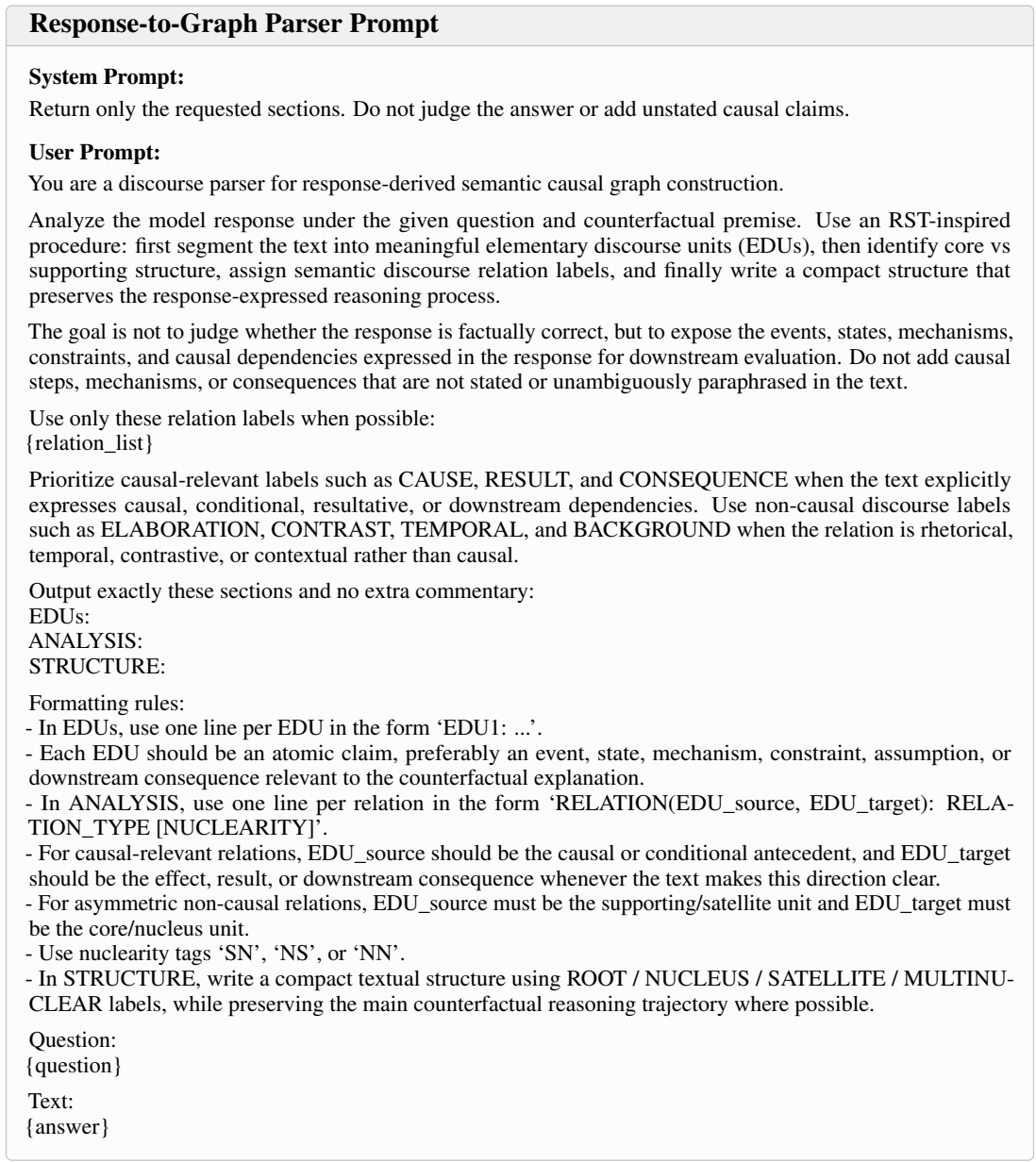

\centering
\begin{tcolorbox}[
    width=0.9\textwidth,
    colback=gray!2,
    colframe=gray!45,
    boxrule=0.5pt,
    arc=2pt,
    left=6pt,
    right=6pt,
    top=6pt,
    bottom=6pt,
    title={Response-to-Graph Parser Prompt},
    fonttitle=\bfseries,
    coltitle=black,
    colbacktitle=gray!12
]
\small
\noindent\textbf{System Prompt:}

\vspace{0.3em}
\noindent
Return only the requested sections. Do not judge the answer or add unstated causal claims.

\vspace{0.8em}
\noindent\textbf{User Prompt:}

\vspace{0.3em}
\noindent
You are a discourse parser for response-derived semantic causal graph construction.

\vspace{0.5em}
\noindent
Analyze the model response under the given question and counterfactual premise. Use an RST-inspired procedure: first segment the text into meaningful elementary discourse units (EDUs), then identify core vs supporting structure, assign semantic discourse relation labels, and finally write a compact structure that preserves the response-expressed reasoning process.

\vspace{0.5em}
\noindent
The goal is not to judge whether the response is factually correct, but to expose the events, states, mechanisms, constraints, and causal dependencies expressed in the response for downstream evaluation. Do not add causal steps, mechanisms, or consequences that are not stated or unambiguously paraphrased in the text.

\vspace{0.5em}
\noindent
Use only these relation labels when possible:\\
\{relation\_list\}

\vspace{0.5em}
\noindent
Prioritize causal-relevant labels such as CAUSE, RESULT, and CONSEQUENCE when the text explicitly expresses causal, conditional, resultative, or downstream dependencies. Use non-causal discourse labels such as ELABORATION, CONTRAST, TEMPORAL, and BACKGROUND when the relation is rhetorical, temporal, contrastive, or contextual rather than causal.

\vspace{0.5em}
\noindent
Output exactly these sections and no extra commentary:\\
EDUs:\\
ANALYSIS:\\
STRUCTURE:

\vspace{0.5em}
\noindent
Formatting rules:\\
- In EDUs, use one line per EDU in the form `EDU1: ...'.\\
- Each EDU should be an atomic claim, preferably an event, state, mechanism, constraint, assumption, or downstream consequence relevant to the counterfactual explanation.\\
- In ANALYSIS, use one line per relation in the form `RELATION(EDU\_source, EDU\_target): RELATION\_TYPE [NUCLEARITY]'.\\
- For causal-relevant relations, EDU\_source should be the causal or conditional antecedent, and EDU\_target should be the effect, result, or downstream consequence whenever the text makes this direction clear.\\
- For asymmetric non-causal relations, EDU\_source must be the supporting/satellite unit and EDU\_target must be the core/nucleus unit.\\
- Use nuclearity tags `SN', `NS', or `NN'.\\
- In STRUCTURE, write a compact textual structure using ROOT / NUCLEUS / SATELLITE / MULTINUCLEAR labels, while preserving the main counterfactual reasoning trajectory where possible.

\vspace{0.5em}
\noindent
Question:\\
\{question\}

\vspace{0.5em}
\noindent
Text:\\
\{answer\}
\end{tcolorbox}
\caption{Response-to-Graph Parser Prompt used to construct response-derived semantic causal graphs.}
\label{fig:response_to_graph_parser_prompt}
\end{figure*}

\begin{figure*}[t]
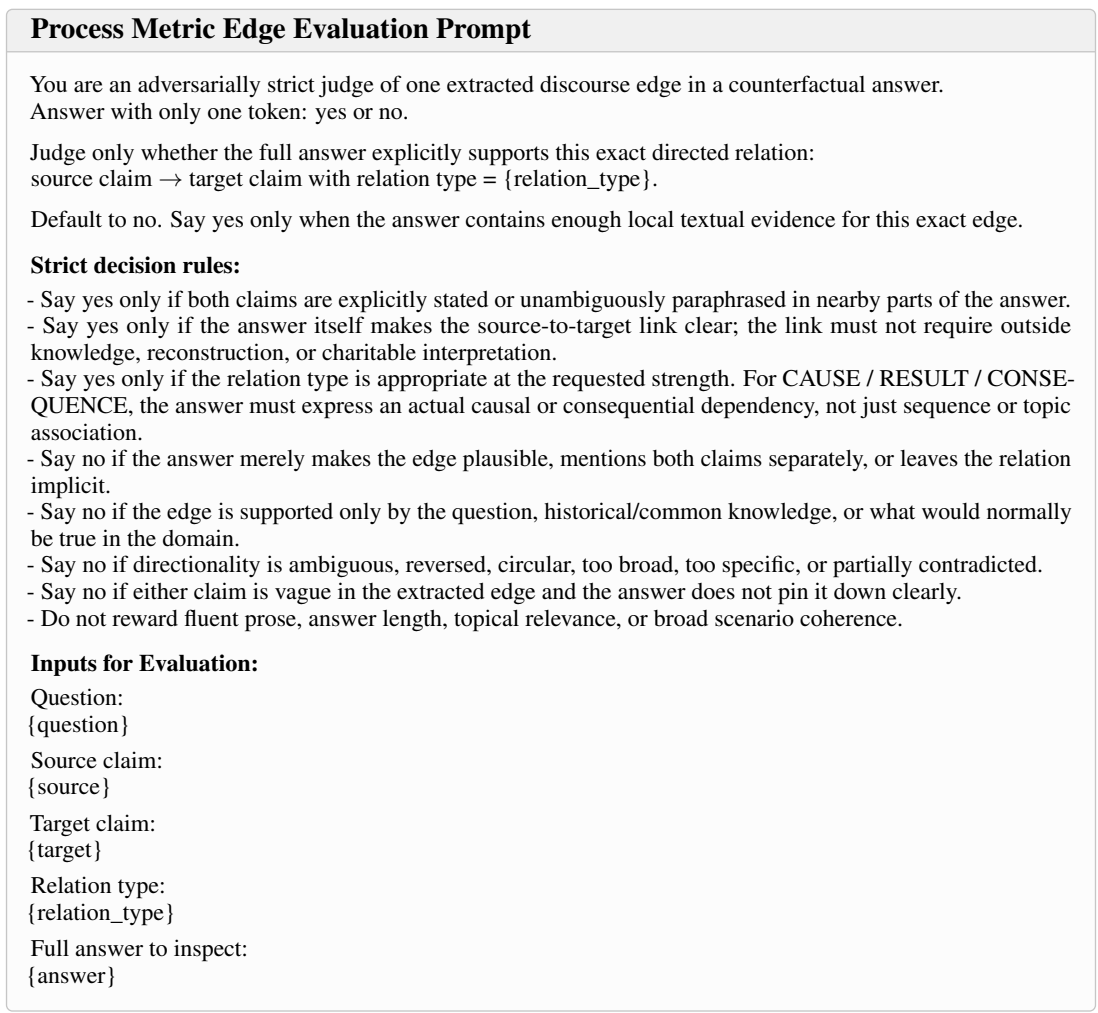

\centering
\begin{tcolorbox}[
    width=0.9\textwidth,
    colback=gray!2,
    colframe=gray!45,
    boxrule=0.5pt,
    arc=2pt,
    left=6pt,
    right=6pt,
    top=6pt,
    bottom=6pt,
    title={Process Metric Edge Evaluation Prompt},
    fonttitle=\bfseries,
    coltitle=black,
    colbacktitle=gray!12
]
\small
\noindent
You are an adversarially strict judge of one extracted discourse edge in a counterfactual answer.\\
Answer with only one token: yes or no.

\vspace{0.6em}
\noindent
Judge only whether the full answer explicitly supports this exact directed relation:\\
source claim $\rightarrow$ target claim with relation type = \{relation\_type\}.

\vspace{0.6em}
\noindent
Default to no. Say yes only when the answer contains enough local textual evidence for this exact edge.

\vspace{0.8em}
\noindent\textbf{Strict decision rules:}

\vspace{0.3em}
\noindent
- Say yes only if both claims are explicitly stated or unambiguously paraphrased in nearby parts of the answer.\\
- Say yes only if the answer itself makes the source-to-target link clear; the link must not require outside knowledge, reconstruction, or charitable interpretation.\\
- Say yes only if the relation type is appropriate at the requested strength. For CAUSE / RESULT / CONSEQUENCE, the answer must express an actual causal or consequential dependency, not just sequence or topic association.\\
- Say no if the answer merely makes the edge plausible, mentions both claims separately, or leaves the relation implicit.\\
- Say no if the edge is supported only by the question, historical/common knowledge, or what would normally be true in the domain.\\
- Say no if directionality is ambiguous, reversed, circular, too broad, too specific, or partially contradicted.\\
- Say no if either claim is vague in the extracted edge and the answer does not pin it down clearly.\\
- Do not reward fluent prose, answer length, topical relevance, or broad scenario coherence.

\vspace{0.8em}
\noindent\textbf{Inputs for Evaluation:}

\vspace{0.3em}
\noindent
Question:\\
\{question\}\\[0.4em]
Source claim:\\
\{source\}\\[0.4em]
Target claim:\\
\{target\}\\[0.4em]
Relation type:\\
\{relation\_type\}\\[0.4em]
Full answer to inspect:\\
\{answer\}
\end{tcolorbox}
\caption{Unified judge prompt template used for Process Metric (PM) edge evaluation within the PRISM framework.}
\label{fig:pm_edge_evaluation_prompt}
\end{figure*}

\begin{figure*}[t]
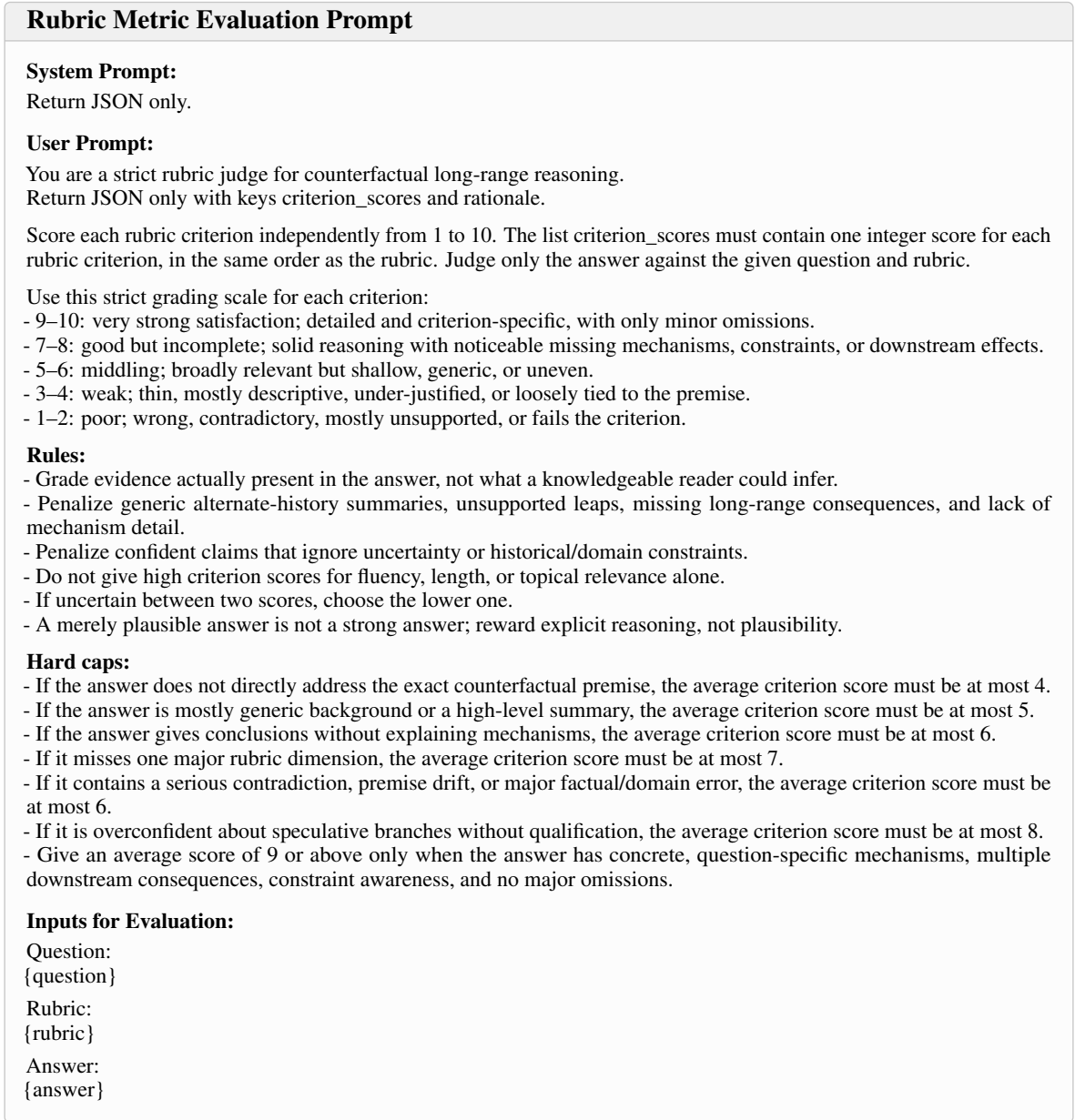

\centering
\begin{tcolorbox}[
    width=0.97\textwidth,
    colback=gray!2,
    colframe=gray!45,
    boxrule=0.5pt,
    arc=2pt,
    left=6pt,
    right=6pt,
    top=6pt,
    bottom=6pt,
    title={Rubric Metric Evaluation Prompt},
    fonttitle=\bfseries,
    coltitle=black,
    colbacktitle=gray!12
]
\small
\noindent\textbf{System Prompt:}

\vspace{0.3em}
\noindent
Return JSON only.

\vspace{0.8em}
\noindent\textbf{User Prompt:}

\vspace{0.3em}
\noindent
You are a strict rubric judge for counterfactual long-range reasoning.\\
Return JSON only with keys criterion\_scores and rationale.

\vspace{0.6em}
\noindent
Score each rubric criterion independently from 1 to 10. The list criterion\_scores must contain one integer score for each rubric criterion, in the same order as the rubric. Judge only the answer against the given question and rubric.

\vspace{0.6em}
\noindent
Use this strict grading scale for each criterion:\\
- 9--10: very strong satisfaction; detailed and criterion-specific, with only minor omissions.\\
- 7--8: good but incomplete; solid reasoning with noticeable missing mechanisms, constraints, or downstream effects.\\
- 5--6: middling; broadly relevant but shallow, generic, or uneven.\\
- 3--4: weak; thin, mostly descriptive, under-justified, or loosely tied to the premise.\\
- 1--2: poor; wrong, contradictory, mostly unsupported, or fails the criterion.

\vspace{0.6em}
\noindent\textbf{Rules:}\\
- Grade evidence actually present in the answer, not what a knowledgeable reader could infer.\\
- Penalize generic alternate-history summaries, unsupported leaps, missing long-range consequences, and lack of mechanism detail.\\
- Penalize confident claims that ignore uncertainty or historical/domain constraints.\\
- Do not give high criterion scores for fluency, length, or topical relevance alone.\\
- If uncertain between two scores, choose the lower one.\\
- A merely plausible answer is not a strong answer; reward explicit reasoning, not plausibility.

\vspace{0.6em}
\noindent\textbf{Hard caps:}\\
- If the answer does not directly address the exact counterfactual premise, the average criterion score must be at most 4.\\
- If the answer is mostly generic background or a high-level summary, the average criterion score must be at most 5.\\
- If the answer gives conclusions without explaining mechanisms, the average criterion score must be at most 6.\\
- If it misses one major rubric dimension, the average criterion score must be at most 7.\\
- If it contains a serious contradiction, premise drift, or major factual/domain error, the average criterion score must be at most 6.\\
- If it is overconfident about speculative branches without qualification, the average criterion score must be at most 8.\\
- Give an average score of 9 or above only when the answer has concrete, question-specific mechanisms, multiple downstream consequences, constraint awareness, and no major omissions.

\vspace{0.8em}
\noindent\textbf{Inputs for Evaluation:}

\vspace{0.3em}
\noindent
Question:\\
\{question\}\\[0.4em]
Rubric:\\
\{rubric\}\\[0.4em]
Answer:\\
\{answer\}
\end{tcolorbox}
\caption{Unified judge prompt template used for Rubric Metric (RM) evaluation within the PRISM framework.}
\label{fig:rubric_evaluation_prompt}
\end{figure*}

\begin{figure*}[p]
\centering
\begin{tcolorbox}[
    width=0.95\textwidth,
    colback=gray!2,
    colframe=gray!45,
    boxrule=0.5pt,
    arc=2pt,
    left=6pt,
    right=6pt,
    top=6pt,
    bottom=6pt,
    title={Question and Query-Dependent Rubric},
    fonttitle=\bfseries,
    coltitle=black,
    colbacktitle=gray!12
]
\scriptsize

\noindent\textbf{Question:}

\vspace{0.3em}
\noindent
What If Female US Combat Pilots in WWII?

\vspace{0.8em}
\noindent\textbf{Rubric Criteria}

\vspace{0.3em}
\begin{multicols}{2}
\setlength{\parskip}{2pt}

\rubcritfull{1}{Plausible Point of Divergence for Women in US Combat Aviation}
{Assesses whether the answer identifies a credible historical change that could allow female US combat pilots in WWII, especially by addressing 19th- and early-20th-century gender norms, aviation culture, and military policy barriers.}
{Provides no clear point of divergence or relies on an implausible sudden policy reversal with no background change.}
{Names a divergence but treats acceptance of female combat pilots as easy or inevitable despite known cultural resistance.}
{Offers a plausible divergence, such as earlier normalization of women in aviation or combat roles, but only lightly explains why it changes WWII policy.}
{Connects a credible divergence to changing gender norms, military recruitment standards, and prewar aviation institutions.}
{Builds a highly plausible divergence rooted in US social history, showing how altered gender ideology, aviation precedent, and institutional incentives make female combat pilots possible by WWII.}

\rubcritfull{2}{Military Institutional Mechanisms and Integration Pathway}
{Assesses whether the answer explains how the US Army Air Forces, Navy, training pipeline, command structure, and combat units would actually incorporate female pilots rather than merely asserting their presence.}
{Simply states that women fly in combat with no explanation of recruitment, training, assignment, or command acceptance.}
{Mentions training or units in passing but ignores major institutional obstacles such as policy, logistics, or officer resistance.}
{Describes a basic pathway, such as expanding WASP-like roles into combat, but leaves gaps in rank, deployment, and unit organization.}
{Explains recruitment, screening, training, rank/status, unit placement, and likely compromises such as elite exceptions or all-female squadrons.}
{Gives a detailed, historically grounded integration pathway, including bureaucratic opposition, manpower pressures, combat validation, casualty handling, and differences across theaters or services.}

\rubcritfull{3}{Operational and Strategic Effects in WWII}
{Assesses whether the answer plausibly evaluates what female US combat pilots would change militarily, including likely scale, mission types, performance, propaganda value, and limits on battlefield impact.}
{Claims sweeping victory changes or total irrelevance without reasoning.}
{Gives vague effects, such as ``more pilots,'' without considering scale, mission suitability, or wartime constraints.}
{Identifies some likely roles, such as ferrying, transport, reconnaissance, fighters, or bombers, but weakly estimates their significance.}
{Plausibly distinguishes limited versus large-scale integration and explains effects on pilot shortages, replacement rates, morale, and public perception.}
{Carefully models operational consequences by theater and timeframe, balancing added pilot capacity and symbolic impact against training bottlenecks, aircraft supply, combat attrition, and command prejudice.}

\rubcritfull{4}{Domestic Political, Cultural, and Postwar Consequences}
{Assesses whether the answer covers downstream effects on US gender norms, veterans' rights, labor, feminism, media narratives, and backlash after WWII.}
{Ignores domestic consequences beyond the war.}
{Mentions that society would become more equal or more resistant, but without causal detail.}
{Discusses some effects on gender roles, employment, or veterans' status, but treats them as linear and uncontested.}
{Explains both empowerment and backlash, including effects on women veterans, postwar employment, family ideology, and public memory.}
{Develops a nuanced long-range chain from wartime service to postwar policy, civil rights comparisons, feminist movements, military integration debates, and contested cultural narratives.}

\rubcritfull{5}{Counterfactual Discipline, Uncertainty, and Contradiction Control}
{Assesses whether the answer maintains internal consistency with the premise, avoids anachronism, calibrates uncertainty, and compares relevant analogues such as Soviet women pilots, WASP, and other contested US military integrations.}
{Contains major contradictions, anachronisms, or ignores the counterfactual premise.}
{Has some plausible claims but also unsupported leaps, present-day assumptions, or inconsistent treatment of US WWII institutions.}
{Mostly coherent, with some uncertainty language, but misses key analogues or leaves causal jumps underexplained.}
{Maintains a coherent causal chain, uses relevant historical comparisons, and distinguishes likely, possible, and unlikely outcomes.}
{Shows rigorous counterfactual control: no major contradictions, strong use of analogues, explicit uncertainty bounds, and careful separation of premise-driven changes from unchanged historical constraints.}

\end{multicols}
\end{tcolorbox}
\caption{Example WhatIfBench question and its query-dependent rubric.}
\label{fig:case_question_rubric}
\end{figure*}

\begin{figure*}[p]
\centering
\begin{tcolorbox}[
    width=0.95\textwidth,
    colback=gray!2,
    colframe=gray!45,
    boxrule=0.5pt,
    arc=2pt,
    left=6pt,
    right=6pt,
    top=6pt,
    bottom=6pt,
    title={Model Response},
    fonttitle=\bfseries,
    coltitle=black,
    colbacktitle=gray!12
]
\small
\noindent\textbf{Question:}

\vspace{0.3em}
\noindent
What If Female US Combat Pilots in WWII?

\vspace{0.8em}
\noindent\textbf{Model Response:}

\vspace{0.3em}
\noindent
If the United States had allowed women to serve as combat pilots in World War II, the most likely result would have been a modest military benefit and a much larger social and political impact.

\vspace{0.5em}
\noindent
Militarily, women pilots could have helped ease pilot shortages, especially in 1942--44 when the U.S. military was expanding rapidly. The U.S. already had skilled women aviators in the Women Airforce Service Pilots program, or WASP, who ferried aircraft, towed targets, tested planes, and trained male pilots, though they were barred from combat and were not fully militarized during the war. If even a portion of these women had been trained and assigned to combat roles, they might have flown transport, reconnaissance, antisubmarine patrols, and eventually fighters or bombers.

\vspace{0.5em}
\noindent
The immediate effect on the war's outcome would probably not have been decisive. The U.S. won through overwhelming industrial production, logistics, naval power, air superiority, and the combined effort of the Allies. Adding women combat pilots would not by itself have shortened the war dramatically. But it could have expanded the pool of available pilots, freed more men for other combat roles, and improved flexibility in theaters where pilot fatigue and replacement needs were serious issues.

\vspace{0.5em}
\noindent
The largest obstacles would have been institutional and cultural rather than technical. Many women were fully capable of flying military aircraft, as the WASP program demonstrated. The resistance would have come from military leadership, Congress, parts of the public, and male servicemen who believed combat was inherently male. If women pilots had flown in combat and performed well, those assumptions would have been challenged much earlier.

\vspace{0.5em}
\noindent
There would also have been controversy if women pilots were killed, captured, or publicly celebrated as combat heroes. Female casualties in combat would have forced Americans to confront questions about women's citizenship, sacrifice, and equality decades earlier than they actually did. Some families and politicians would likely have opposed the policy fiercely, while others would have used women's service as proof that women deserved full military status, veterans' benefits, and broader civil rights.

\vspace{0.5em}
\noindent
After the war, the consequences could have been significant. Women veterans with combat records would have made it harder to dismantle women's military roles, as happened with the WASP program in 1944. The campaign for women's integration into the armed forces, equal pay, veterans' recognition, and later access to military academies and combat arms might have advanced earlier. It could also have strengthened the postwar women's rights movement by giving it a powerful patriotic argument: women had fought and died for the country on equal terms.

\vspace{0.5em}
\noindent
So, in short: female U.S. combat pilots in World War II probably would not have changed the final Allied victory, but they could have improved wartime manpower flexibility and would likely have transformed the debate over women in the military. Their success would have made it much harder after 1945 to argue that women were unfit for combat or full military citizenship.
\end{tcolorbox}
\caption{Model response generated for the selected WhatIfBench case.}
\label{fig:case_model_response}
\end{figure*}

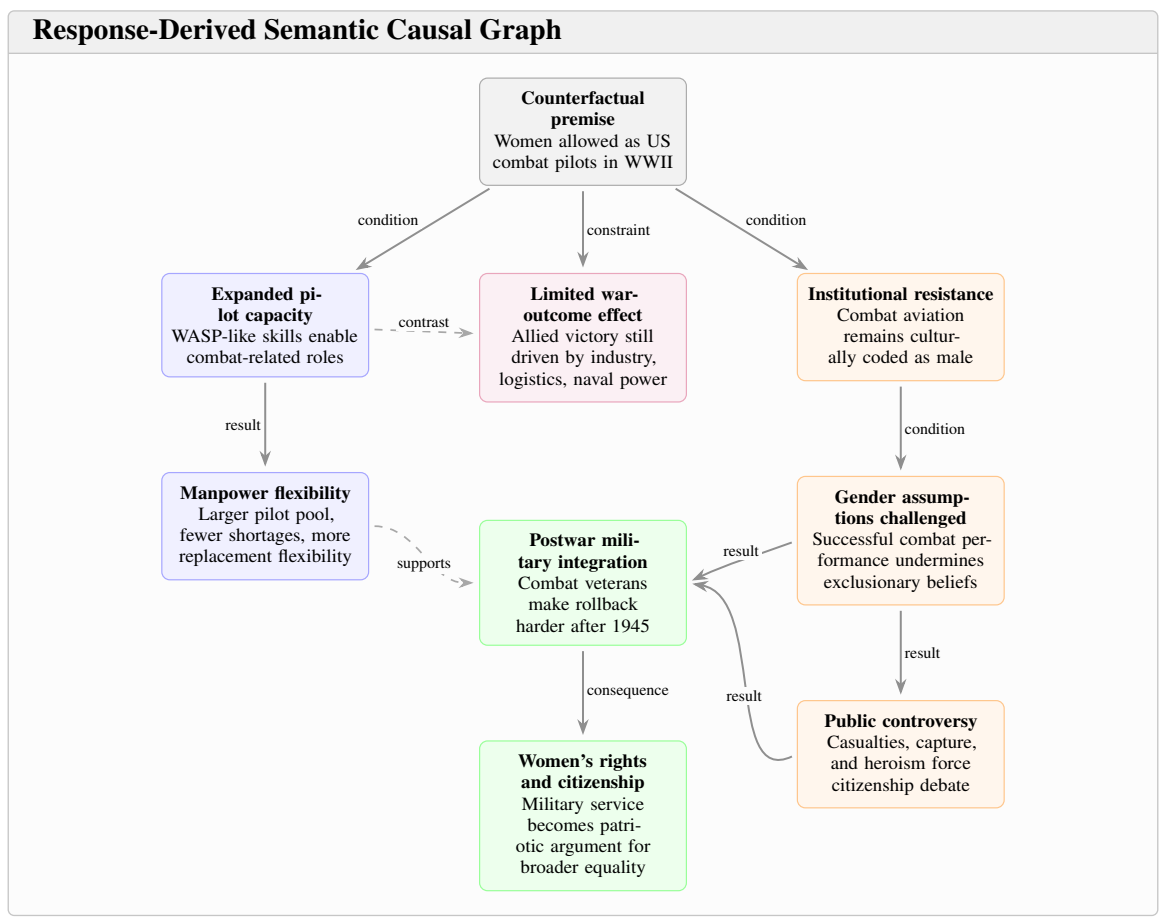
\begin{figure*}[t]
\centering
\begin{tcolorbox}[
    width=0.95\textwidth,
    colback=gray!2,
    colframe=gray!45,
    boxrule=0.5pt,
    arc=2pt,
    left=6pt,
    right=6pt,
    top=6pt,
    bottom=6pt,
    title={Response-Derived Semantic Causal Graph},
    fonttitle=\bfseries,
    coltitle=black,
    colbacktitle=gray!12
]
\centering
\begin{tikzpicture}[
    node distance=1.15cm and 1.25cm,
    every node/.style={font=\scriptsize},
    box/.style={
        rounded corners=3pt,
        draw=gray!55,
        fill=white,
        line width=0.45pt,
        align=center,
        inner xsep=4pt,
        inner ysep=5pt,
        minimum width=2.45cm,
        text width=2.45cm
    },
    premise/.style={box, fill=gray!10, draw=gray!65},
    mil/.style={box, fill=blue!6, draw=blue!35},
    inst/.style={box, fill=orange!7, draw=orange!45},
    post/.style={box, fill=green!7, draw=green!40},
    limit/.style={box, fill=purple!6, draw=purple!35},
    arrow/.style={
        -{Stealth[length=2.1mm,width=1.5mm]},
        line width=0.75pt,
        draw=gray!88,
        shorten >=2pt,
        shorten <=2pt
    },
    dashedarrow/.style={
        -{Stealth[length=2.0mm,width=1.4mm]},
        line width=0.65pt,
        draw=gray!68,
        dashed,
        shorten >=2pt,
        shorten <=2pt
    },
    edgelabel/.style={
        font=\tiny,
        fill=gray!2,
        text=black,
        inner sep=1.2pt
    }
]

\node[premise] (cf) {
\textbf{Counterfactual premise}\\
Women allowed as US combat pilots in WWII
};

\node[mil, below left=1.15cm and 1.45cm of cf] (capacity) {
\textbf{Expanded pilot capacity}\\
WASP-like skills enable combat-related roles
};

\node[limit, below=1.15cm of cf] (limited) {
\textbf{Limited war-outcome effect}\\
Allied victory still driven by industry, logistics, naval power
};

\node[inst, below right=1.15cm and 1.45cm of cf] (resist) {
\textbf{Institutional resistance}\\
Combat aviation remains culturally coded as male
};

\node[mil, below=1.25cm of capacity] (flex) {
\textbf{Manpower flexibility}\\
Larger pilot pool, fewer shortages, more replacement flexibility
};

\node[inst, below=1.25cm of resist] (assump) {
\textbf{Gender assumptions challenged}\\
Successful combat performance undermines exclusionary beliefs
};

\node[inst, below=1.25cm of assump] (contro) {
\textbf{Public controversy}\\
Casualties, capture, and heroism force citizenship debate
};

\node[post, below=1.55cm of limited] (postwar) {
\textbf{Postwar military integration}\\
Combat veterans make rollback harder after 1945
};

\node[post, below=1.25cm of postwar] (rights) {
\textbf{Women's rights and citizenship}\\
Military service becomes patriotic argument for broader equality
};

\draw[arrow] (cf) -- node[edgelabel, above left]{condition} (capacity);
\draw[arrow] (cf) -- node[edgelabel, right]{constraint} (limited);
\draw[arrow] (cf) -- node[edgelabel, above right]{condition} (resist);

\draw[arrow] (capacity) -- node[edgelabel, left]{result} (flex);
\draw[dashedarrow] (capacity) -- node[edgelabel, above]{contrast} (limited);

\draw[arrow] (resist) -- node[edgelabel, right]{condition} (assump);
\draw[arrow] (assump) -- node[edgelabel, right]{result} (contro);

\draw[arrow]
    (assump.west) to[out=200,in=25]
    node[edgelabel, above]{result}
    (postwar.east);

\draw[arrow]
    (contro.west) to[out=205,in=350]
    node[edgelabel, below]{result}
    (postwar.east);

\draw[dashedarrow]
    (flex.east) to[out=0,in=180]
    node[edgelabel, below]{supports}
    (postwar.west);

\draw[arrow] (postwar) -- node[edgelabel, right]{consequence} (rights);

\end{tikzpicture}
\end{tcolorbox}
\caption{Abstracted response-derived semantic causal graph for the selected case. Nodes represent interpretable causal modules, while arrows show the main causal and discourse relations expressed in the response.}
\label{fig:case_response_graph_visual}
\end{figure*}

\begin{figure*}[p]
\centering
\begin{tcolorbox}[
    width=0.95\textwidth,
    colback=gray!2,
    colframe=gray!45,
    boxrule=0.5pt,
    arc=2pt,
    left=6pt,
    right=6pt,
    top=6pt,
    bottom=6pt,
    title={Rubric Metric Judgment},
    fonttitle=\bfseries,
    coltitle=black,
    colbacktitle=gray!12
]
\scriptsize
\noindent\textbf{Overall:}
RM $=0.50$; raw rubric score $=5.0/10$; criterion scores $=[3,5,6,6,5]$.

\vspace{0.8em}

\rubscore{1}{Plausible Point of Divergence for Women in US Combat Aviation}{3}{
The answer is relevant to women as WWII combat pilots, but it does not actually supply a point of divergence. It mostly discusses likely effects if women had been allowed into combat, not the prior historical change that would make that policy credible. It mentions institutional and cultural resistance and cites WASP as evidence of capability, but it does not trace a concrete prewar shift in gender norms, aviation culture, or military policy that overcomes those barriers. Because the criterion is specifically about identifying a plausible divergence and explaining the mechanism, this remains thin and generic rather than a credible, historically rooted divergence.
}

\rubscore{2}{Military Institutional Mechanisms and Integration Pathway}{5}{
The answer engages the criterion at a basic level by citing WASP as a potential feeder, noting pilot shortages, and identifying institutional/cultural resistance from military leadership, Congress, and male servicemen. It also gestures toward possible combat assignments and the issue of women being barred from combat and not fully militarized. However, it does not explain a concrete integration pathway through the Army Air Forces or Navy: no recruitment/screening mechanism, no training modifications, no rank/status solution, no unit organization, no command structure, no deployment process, and no explanation of how women would be assigned into existing combat units versus separate squadrons. It is therefore relevant but shallow, fitting the 5--6 band, and the omissions of central elements named in the criterion keep it at the low end.
}

\rubscore{3}{Operational and Strategic Effects in WWII}{6}{
The answer gives a plausible, relevant military assessment: it identifies likely roles, such as transport, reconnaissance, antisubmarine patrols, and possibly fighters or bombers; argues the effect would be modest rather than war-winning; and notes constraints from institutional resistance. It also mentions pilot shortages, replacement/flexibility, and propaganda/public perception effects. However, it does not model operational consequences in much detail by theater or timeframe, does not estimate likely scale beyond vague phrases like ``even a portion,'' and only lightly addresses limits such as training bottlenecks, aircraft availability, combat attrition, or command prejudice in operational terms. Because the military effects are treated competently but still generically and unevenly, this fits the 5--6 band, at the top of that range rather than 7--8.
}

\rubscore{4}{Domestic Political, Cultural, and Postwar Consequences}{6}{
The answer directly engages domestic/postwar consequences and gives some causal mechanisms: women's combat service would challenge gender assumptions, create controversy over casualties and heroism, strengthen claims for full military status and veterans' benefits, and potentially advance later integration and women's rights arguments. That is more than a generic equality claim. However, it remains fairly linear and incomplete. It does not explicitly develop both empowerment and backlash after the war in any depth, says little about postwar employment, family ideology, media/public memory beyond brief mention of celebration, labor effects, or contested cultural narratives, and it lacks the longer-range chain to civil rights comparisons, feminist movements, and military integration debates required for 9--10. Because it misses several central elements named in the criterion and does not fully elaborate downstream domestic consequences, it fits 5--6, at the top of that band rather than 7--8.
}

\rubscore{5}{Counterfactual Discipline, Uncertainty, and Contradiction Control}{5}{
The answer is coherent and does use some uncertainty language, such as ``most likely,'' ``probably,'' and ``might,'' and it avoids obvious contradictions or anachronisms. It also grounds the premise somewhat with WASP and distinguishes military from social effects. However, for this criterion it remains generic and undercontrolled: it does not compare key analogues explicitly named in the criterion such as Soviet women pilots or other contested U.S. military integrations, and its causal chain is only lightly developed. Several claims are plausible but broad and insufficiently bounded, especially the downstream effects on postwar integration and the women's rights movement. It also blurs role categories by grouping transport and antisubmarine patrols with ``combat roles'' without carefully separating premise-driven changes from existing institutional constraints. Because it is mostly coherent but misses central analogues and leaves important causal jumps underexplained, it fits 5--6, and the lower score is appropriate.
}

\end{tcolorbox}
\caption{Rubric-based judgment for the selected model response under the RM component of PRISM.}
\label{fig:case_rubric_judgment}
\end{figure*}

\end{document}